\documentclass[journal=jacsat,manuscript=article]{achemso}

\usepackage[version=3]{mhchem} % Formula subscripts using \ce{}
\SectionNumbersOn   
\usepackage{multirow}
\usepackage[table]{xcolor}
\usepackage{booktabs}
\usepackage{amsfonts}
\usepackage{amsmath}
\usepackage{pdfpages}

\usepackage{tikz}
\usetikzlibrary{arrows.meta,positioning,fit,calc}
\tikzset{>=Stealth}

\usepackage{algorithm}
\usepackage{algpseudocode}
\usepackage{setspace}

\usepackage{xcolor}
\usepackage[normalem]{ulem} % enable \sout for strikeout

\author{Xu Han}
\affiliation[University]{University of Chinese Academy of Sciences, Beijing, 100190, China}
\alsoaffiliation[Laboratory]{Key Laboratory of Complex Systems Cognition and Decision, Institute of Automation, Chinese Academy of Sciences, Beijing, 100190, China}
\alsoaffiliation[Company]{Beijing Academy of Artificial Intelligence, Beijing, 100083, China}

\author{Yuancheng Sun}
\affiliation[University]{University of Chinese Academy of Sciences, Beijing, 100190, China}
\alsoaffiliation[Laboratory]{Key Laboratory of Complex Systems Cognition and Decision, Institute of Automation, Chinese Academy of Sciences, Beijing, 100190, China}
\alsoaffiliation[Company]{Beijing Academy of Artificial Intelligence, Beijing, 100083, China}

\author{Kai Chen}
\affiliation[Company]{Beijing Academy of Artificial Intelligence, Beijing, 100083, China}

\author{Yuxuan Ren}
\affiliation[Company]{Beijing Academy of Artificial Intelligence, Beijing, 100083, China}

\author{Kang Liu}
\affiliation[University]{University of Chinese Academy of Sciences, Beijing, 100190, China}
\alsoaffiliation[Laboratory]{Key Laboratory of Complex Systems Cognition and Decision, Institute of Automation, Chinese Academy of Sciences, Beijing, 100190, China}
\alsoaffiliation[Company]{Beijing Academy of Artificial Intelligence, Beijing, 100083, China}

\author{Qiwei Ye}
\affiliation[Company]{Beijing Academy of Artificial Intelligence, Beijing, 100083, China}
\email{qiwei.ye@baai.ac.cn}

\title[Article Title]{Constraint Decoupled Latent Diffusion for Protein Backmapping}

\abbreviations{IR,NMR,UV}

\begin{document}

\clearpage
\newpage

\begin{abstract}
\noindent Coarse-grained (CG) molecular dynamics simulations enable efficient exploration of protein conformational ensembles. 
However, reconstructing atomic details from CG structures (backmapping) remains a challenging problem.
Current approaches face an inherent trade-off between maintaining atomistic accuracy and exploring diverse conformations, often necessitating complex constraint handling or extensive refinement steps.
To address these challenges, we introduce a novel two-stage framework, named CODLAD (COnstraint Decoupled LAtent Diffusion). 
This framework first compresses atomic structures into discrete latent representations, explicitly embedding structural constraints, thereby decoupling constraint handling from generation. 
Subsequently, it performs efficient denoising diffusion in this latent space to produce structurally valid and diverse all-atom conformations. 
Comprehensive evaluations on diverse protein datasets demonstrate that CODLAD achieves state-of-the-art performance in atomistic accuracy, conformational diversity, and computational efficiency while exhibiting strong generalization across different protein systems. Code is available at \url{https://github.com/xiaoxiaokuye/CODLAD}.
\end{abstract}

%%%%%%%%%%%%%%%%%
%%%%%%%%%%%%%%%%%
% Introduction
%%%%%%%%%%%%%%%%%
%%%%%%%%%%%%%%%%%
\section{Introduction}
\label{introduction}

Coarse-grained (CG) force fields streamline molecular dynamics (MD) simulations by representing groups of atoms as unified beads. 
It reduces system complexity and smooths the energy landscape~\citep{souza2021martini,majewski2023machine,arts2023two}. 
This simplified approach enables exploration of long-timescale phenomena, including protein folding and conformational transitions~\citep{lequieu20191cpn,meller2023accelerating}. 
However, CG representations inherently sacrifice atomic-level details, which are crucial for essential tasks such as molecular recognition, protein-ligand docking, and protein-protein interactions~\citep{badaczewska2020computational,vickery2021cg2at2,zambaldi2024novo,jones24flowback}. 

% backmapping
%%
\begin{figure}[t!]
    \centering
    \includegraphics[width=0.7\columnwidth]{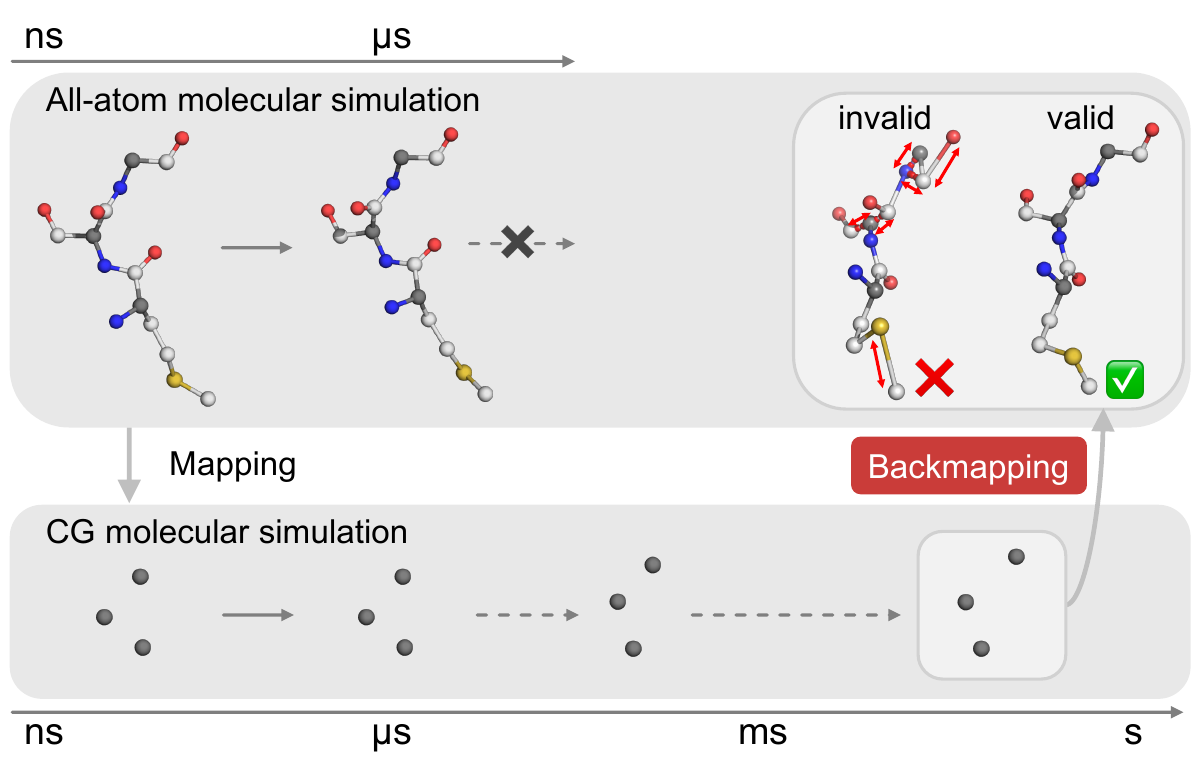}
    \caption{Overview of backmapping. All-atom simulations are limited by high computational cost and short timescales (ns to $\mu$s), while CG simulations can reach longer timescales (ms to s) at lower cost. Backmapping enables recovery of valid atomic details from CG structures and facilitate atomic-level tasks.}
    \label{fig:backmapping}
\end{figure}

Backmapping, the task of reconstructing all-atom structures from CG representations, bridges this gap by restoring the atomic resolution necessary for detailed structural analysis~\citep{lombardi2016cg2aa}. 
As illustrated in Figure~\ref{fig:backmapping}, although CG MD accesses longer timescales (ms to s) that are inaccessible to MD of all atoms, atomic details remain necessary for subsequent structural analysis.
To address this gap, effective backmapping approaches require \textit{atomistic accuracy} in reconstructing physically realistic atomic details and \textit{conformational diversity} to capture the broad spectrum of biologically relevant structural ensembles. 
In addition, they should achieve \textit{computational efficiency} and strong \textit{generalizability} across diverse molecular systems to support scalable and robust application~\citep{luo2022autoregressive,qiang2023coarse}.

Traditional backmapping approaches often generated initial atomic structures using heuristic rule-based algorithms, followed by additional refinement steps, such as geometric optimization or energy minimization~\citep{lombardi2016cg2aa,nicholson2020constructing,vickery2021cg2at2}. 
However, these refinements are computationally expensive~\citep{jones2023diamondback}, frequently compromise structural validity by introducing non-physical artifacts such as atomic clashes~\citep{xu2019opus}, and struggle to capture the diverse conformational states due to their deterministic nature~\citep{wang2022generative}. 
Although generative models such as VAE and GAN offer computational advantages for backmapping, they often struggle to capture the full thermodynamic diversity of molecular systems, resulting in limited chemical transferability and mode collapse~\citep{wang2019coarse,li2020backmapping,stieffenhofer2020adversarial,shmilovich2022temporally,wang2022generative,heo2023one,yang2023chemically,chennakesavalu2024data}.

Recent approaches have explored diffusion models for backmapping, leveraging their promising capabilities in generating diverse and detailed structures through stochastic sampling. 
Despite these advances, current diffusion model applications in backmapping—such as DiAMoNDBack~\citep{jones2023diamondback} and FlowBack~\citep{jones24flowback}—still exhibit certain challenges.
These include increased computational overhead and impaired global geometry due to residue-wise denoising~\citep{jones2023diamondback}, or limited generalization across varied conformational systems~\citep{jones24flowback}.
While latent diffusion has shown promise in protein modeling, prior VAE-based approaches typically focus on backbone-level or short sequences. 
Moreover, these methods are limited in addressing the specific constraints and multi-resolution challenges posed by coarse-grained to all-atom backmapping~\citep{fu2024latent, gao2025foldtoken,hayes2025simulating, wang2024dplm,wang2025elucidating}.

To address these limitations, we propose CODLAD (\textbf{CO}nstraint-\textbf{D}ecoupled \textbf{LA}tent \textbf{D}iffusion), a two-stage framework supporting all-atom modeling and tailored for scalable restoration from low-resolution inputs.
First, CODLAD employs a compression stage that encodes atomic structures into discrete latent features using hierarchical feature extraction. 
This process learns to reconstruct the original structures while preserving their inherent constraints, and the adoption of discrete features helps mitigate mode collapse in the conformational space~\citep{yang2023chemically}. 
Subsequently, this compression decouples explicit structural constraint handling from the generation stage by embedding such information within the latent space. 
Second, the generation stage employs denoising diffusion in this latent space, naturally preserving structural validity throughout the sampling process.
By decoupling constraint handling from generation and operating in a low-dimensional, smoothed latent space, CODLAD produces accurate and diverse reconstructions without requiring explicit structural handling, while also achieving high efficiency and strong generalization across protein systems.

We conduct comprehensive experiments on a range of protein datasets, including the PED benchmark~\citep{ghafouri2024ped}, the ATLAS protein dynamics dataset~\citep{vander2024atlas}, the static PDB dataset~\citep{berman2000protein}, and the fast-folding trajectory DES dataset~\citep{lindorff2011fast}.  
The results demonstrate that CODLAD not only achieves state-of-the-art performance in atomistic accuracy and conformational diversity, but also delivers substantial improvements in inference efficiency over existing methods.  
Furthermore, CODLAD generalizes well across dynamic conformational systems, as demonstrated by its strong performance on the unseen DES dataset.

% main
%%
\begin{figure*}[t!]
    \centering
    \includegraphics[width=1\textwidth]{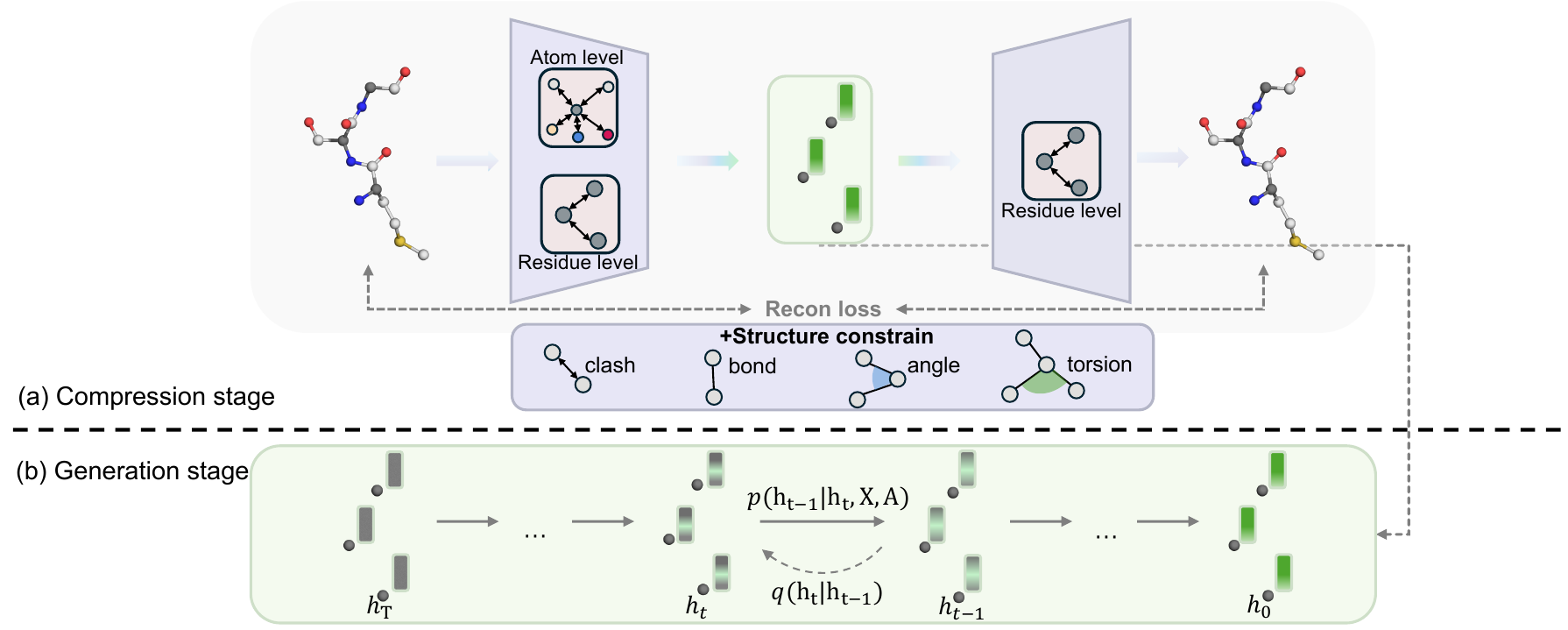} 
    \caption{
    Overview of CODLAD's two-stage framework. (a) Compression stage: All-atom structures are encoded with atom-level and residue-level (C$_\alpha$) message passing, with cross-level information exchange (see Eq.~(\ref{eq:hierarchical_mp})), to produce residue-level latent representations.
    The decoder predicts internal coordinates and deterministically maps them to Cartesian coordinates.
    Reconstruction employs both coordinate-based and structural constraint losses (see Eq.~(\ref{eq:vaeloss})), 
    such as bond-length, bond-angle, and torsion terms, along with a clash penalty to discourage atom overlap.
    (\emph{Purple} denotes atom-level operations; \emph{green} denotes latent modules.) 
    (b) Latent stage:
    From the residue-level latent $\mathbf{h_0}$, a forward noising process $q(\mathbf{h}_t\!\mid\!\mathbf{h}_{t-1})$ gradually adds small Gaussian noise (visualized as \emph{green} $\rightarrow$ \emph{gray} blocks, \emph{Gray} indicate noisy latent states).
    A learned reverse process $p(\mathbf{h}_{t-1}\!\mid\!\mathbf{h}_t,\mathbf{X},\mathbf{A})$ denoises in latent space (conditioned on the CG graph, where $\mathbf{X}$ and $\mathbf{A}$ denote CG coordinates and residue type) to recover valid latent representations, decoded as in (a) to generate all-atom structures following $p(\mathbf{x}\mid\mathbf{X},\mathbf{A})$.
    Panel (a) corresponds to the hierarchical message passing in Eq.~(\ref{eq:hierarchical_mp}) and the autoencoder loss in Eq.~(\ref{eq:vaeloss}); panel (b) follows the conditional denoising objective in Eq.~(\ref{eq:final_denoise_loss}) and the sampling procedure summarized in Algs.~S3–S4.
    }
    \label{fig:mainframe}
\end{figure*}

In summary, the key contributions of this paper are: 
\begin{itemize}
   \item This paper proposes CODLAD, a novel two-stage latent diffusion framework for protein backmapping that decouples structural constraints from the generation process, enabling accurate and diverse reconstruction of atomic structures from CG representations \textit{(see Sec.~\ref{sec:mainresults}} and \textit{Sec.~\ref{sec:ablation})}.
   
   \item A hierarchical compression strategy is designed to transform atomic-level data into discrete latent representations, preserving structural constraints and mitigating mode collapse, thus improving distributional matching and computational efficiency \textit{(see Sec.~\ref{sec:time}} and \textit{Sec.~\ref{sec:dis})}.
    
   \item Comprehensive evaluation on multiple datasets demonstrates that CODLAD achieves state-of-the-art performance by delivering superior atomistic accuracy (e.g., a 30.6\% reduction in Graph Edit Distance), exceptional computational efficiency (a 71.1\% average speedup), and, most notably, strong generalization—providing the first evidence of success on an out-of-distribution system with error reductions up to 56.9\% while maintaining high conformational diversity (see \textit{Sec.~\ref{sec:mainresults}} and \textit{Sec.~\ref{sec:generalization_exp}}).

\end{itemize}

%%%%%%%%%%%%%%%%%
%%%%%%%%%%%%%%%%%
% Method
%%%%%%%%%%%%%%%%%
%%%%%%%%%%%%%%%%%
\section{Method}
\label{method}

\subsection{Preliminaries}
\label{preliminaries}

%%%%%%%%%%%%%%%%%
%%%%%%%%%%%%%%%%%
% Background
%%%%%%%%%%%%%%%%%
%%%%%%%%%%%%%%%%%

\paragraph{Problem Formulation.} 
Protein backmapping reconstructs all-atom (AA) structures from their CG representations. An all-atom structure is represented as $\mathbf{AA} = \left\{ (x_i, a_i) \right\}_{i=1}^{n}$, where $x_i$ denotes atomic coordinates and $a_i$ represents atomic types. 
Similarly, a CG structure is represented as $\mathbf{CG} = \left\{ (X_i, A_i) \right\}_{i=1}^{N}$, where $X_i$ denotes bead coordinates and $A_i$ represents amino acid types. 
Given a CG structure, our goal is to generate atomic coordinates $x$, where atomic types $a$ are determined by the amino acid sequence. 
This task can be formulated as learning to sample from the conditional distribution $p(x \mid X, A)$.

Following established methods~\citep{badaczewska2020computational, yang2023chemically, jones2023diamondback,liu2023backdiff,jones24flowback,zhang2025exploit}, we adopt $C_{\alpha}$ atoms as CG representation, which provides a robust encoding of protein-protein interactions and serves as a reliable granularity for reverse mapping.

\paragraph{Internal Coordinates.}
Internal coordinates are used to represent all-atom protein structures, encoding molecular geometry in terms of bond lengths $d_i$, bond angles $\theta_i$, and dihedral angles $\tau_i$ rather than absolute Cartesian positions.
This representation, denoted as $\mathcal{T} = \{(d_i, \theta_i, \tau_i)\}_{i=1}^{K}$, where $K$ is the total number of triplets (typically $N \times 13$ for $N$ residues, assuming up to 13 heavy atoms per residue), offers improved numerical stability for flexible molecular systems~\citep{jing2022torsional, yang2023chemically}.  
Since internal coordinates capture \emph{relative} spatial relationships between atoms, they preserve essential structural features while allowing uniform processing across residues.  
Residues with fewer than 13 heavy atoms are padded with virtual atoms to maintain a fixed representation size for computational models. The internal coordinates are defined as follows.

(1) Bond Lengths:
Bond lengths represent the distances between two bonded atoms. For two atoms $ i $ and $ j $, the bond length $ d_{ij} $ is: $d_{ij} = \|\mathbf{x}_i - \mathbf{x}_j\|,$ where $ \mathbf{x}_i $ and $ \mathbf{x}_j $ are their Cartesian coordinates.

(2) Bond Angles:
Bond angles describe the angles formed by three consecutive atoms. For atoms $ i $, $ j $, and $ k $, the bond angle $ \theta_{ijk} $ is calculated as:
\begin{equation}
\theta_{ijk} = \arccos\left(\frac{(\mathbf{x}_i - \mathbf{x}_j) \cdot (\mathbf{x}_k - \mathbf{x}_j)}{\|\mathbf{x}_i - \mathbf{x}_j\| \|\mathbf{x}_k - \mathbf{x}_j\|}\right),
\end{equation}
ensuring the spatial orientation of bonded atoms.

(3) Dihedral Angles:  
Dihedral angles measure the rotation around a bond and are defined by four consecutive atoms. For atoms $ i $, $ j $, $ k $, and $ l $, the dihedral angle $ \tau_{ijkl} $ is:
\begin{equation}
\tau_{ijkl} = \arctan2\left(\frac{(\mathbf{b}_1 \times \mathbf{b}_2) \cdot \mathbf{b}_3}{\|\mathbf{b}_2\| \mathbf{b}_1 \cdot \mathbf{b}_3}, (\mathbf{b}_1 \times \mathbf{b}_2) \cdot (\mathbf{b}_2 \times \mathbf{b}_3)\right),
\end{equation}

where: $\mathbf{b}_1 = \mathbf{x}_j - \mathbf{x}_i, \quad 
\mathbf{b}_2 = \mathbf{x}_k - \mathbf{x}_j, \quad 
\mathbf{b}_3 = \mathbf{x}_l - \mathbf{x}_k$.
Dihedral angles are critical for capturing the rotational flexibility of residues, particularly in side chains.

The methodology converts protein structures into internal coordinates in two stages, following a predefined atom order.
Backbone atoms are processed first, using $\mathrm{C}_\alpha$ atoms from adjacent residues to establish the structural framework.
Side chain atoms are then added sequentially, starting from known backbone atoms ($\mathrm{N}$, $\mathrm{C}_\alpha$, $\mathrm{C}$) as references for bond lengths, bond angles, and dihedral angles.

\begin{figure}[htbp]
    \centering
    \includegraphics[width=0.3\textwidth]
    {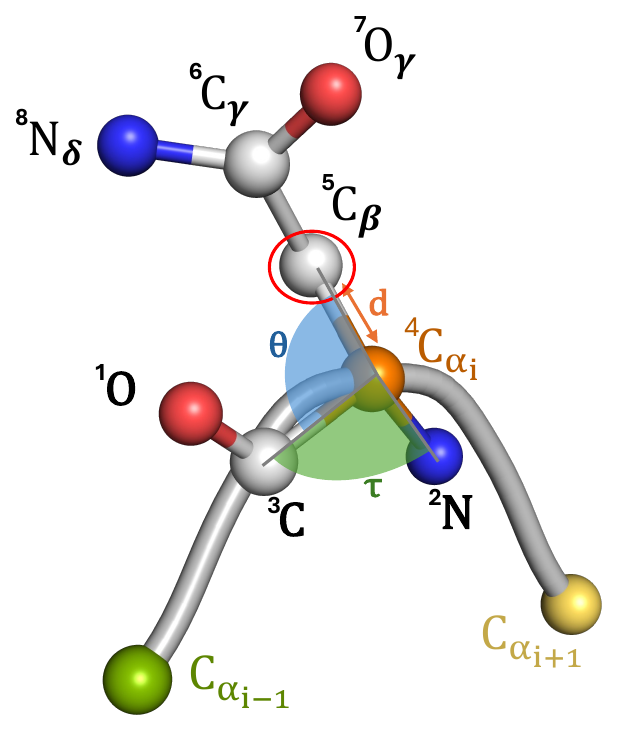} 
    \caption{Schematic representation of internal coordinates in a protein residue. For illustrative purposes, the conversion process is demonstrated using the $\text{C}_\beta$ atom (labeled as atom 5) as an example: (1) The bond length $ d $ is computed as the distance between $\text{C}_\beta$ (atom 5) and $\text{C}$ (atom 4). (2) The bond angle $ \theta $ is calculated as the angle formed by $\text{N}$ (atom 3), $\text{C}$ (atom 4), and $\text{C}_\beta$ (atom 5). (3) The dihedral angle $ \tau $ is determined from the planes formed by $\text{C}_\alpha$ (atom 2), $\text{N}$ (atom 3), $\text{C}$ (atom 4), and $\text{C}_\beta$ (atom 5).}
    \label{fig:ic}
\end{figure}

Figure~\ref{fig:ic} provides a schematic representation of the backbone and side chain atoms of a residue, highlighting the internal coordinate framework. This systematic process ensures that all atoms, including both backbone and side chain atoms, are represented in a consistent and compact internal coordinate framework.

\paragraph{Diffusion Background.} 

Denoising diffusion probabilistic models (DDPMs)~\citep{sohl2015deep, ho2020denoising} operate through two phases: forward diffusion that transforms data into noise, and a reverse process that reconstructs the original data. 
During forward diffusion, the initial data $x_0$ is progressively corrupted into Gaussian noise $x_T$ through Markov transitions. 
The reverse process reconstructs $x_0$ through iterative denoising of states $\{x_t\}_{t=1}^T$. A neural network $\epsilon_\theta(x_t, t)$ predicts noise at each timestep.

The reverse denoising process follows the equation:
\begin{equation}
x_{t-1} = \frac{1}{\sqrt{\alpha_t}} \left( x_t - \frac{1-\alpha_t}{\sqrt{1-\bar{\alpha}_t}} \epsilon_\theta(x_t, t) \right) + \sigma_t z
\end{equation}
where $\alpha_t$ and $\bar{\alpha}_t = \prod_{i=1}^t \alpha_i$ define the noise schedule, $z \sim \mathcal{N}(0, I)$ introduces stochastic variation, and $\sigma_t$ controls its magnitude.

Due to its stochastic nature, this framework is particularly effective for exploring diverse conformations in protein structure generation~\citep{campbell2024generative,chu2024all}. 
To improve efficiency, the operation can be extended into a compressed latent space, a strategy that lowers computational costs~\citep{rombach2022high}. 
Despite this, applying the model to large proteins remains a considerable challenge requiring advanced training methods~\citep{winter2022unsupervised,xu2023geometric,fu2024latent,van2024fast,gao2025foldtoken,hayes2025simulating}.

\subsection{CODLAD Pipeline}
\label{sec:pipeline}

\textbf{Pipeline overview.}
Figure~\ref{fig:mainframe} summarizes the overall workflow of CODLAD, which integrates a hierarchical encoder–decoder and a latent diffusion model to achieve CG-to-AA backmapping. 
The framework operates in two coupled stages: Stage (a) learns residue-level latent representations of all-atom structures that satisfy geometric constraints through hierarchical encoding and reconstruction, while Stage (b) learns a conditional diffusion process in this latent space to generate diverse and physically plausible all-atom conformations conditioned on CG inputs.

\textit{Training.}
In Stage (a), an SE(3)-equivariant hierarchical GNN encodes AA structures into a compact latent space and reconstructs them while enforcing geometric consistency; cross-level message passing between atom and residue graphs is given in Eq.~(\ref{eq:hierarchical_mp}). 
Vector quantization regularizes the latent manifold and yields discrete codes, which can help mitigate mode collapse.
In Stage (b), a diffusion model is trained in this latent space to map noisy latents to clean, constraint-consistent latents conditioned on the CG graph; the objective is Eq.~(\ref{eq:final_denoise_loss}). 
Stage (a) establishes the latent manifold and decoder, while Stage (b) learns CG-conditioned sampling on this manifold.

\textit{Inference.}
Given a new CG input, CODLAD first runs Stage (b) to denoise from a random latent to a clean latent conditioned on the CG graph, and then applies the decoder from Stage (a) to obtain internal coordinates, which are deterministically converted to Cartesian coordinates to produce the final AA structure. 
No encoder is used at inference.

\begin{figure}[t]
    \centering
    \includegraphics[width=0.6\columnwidth]{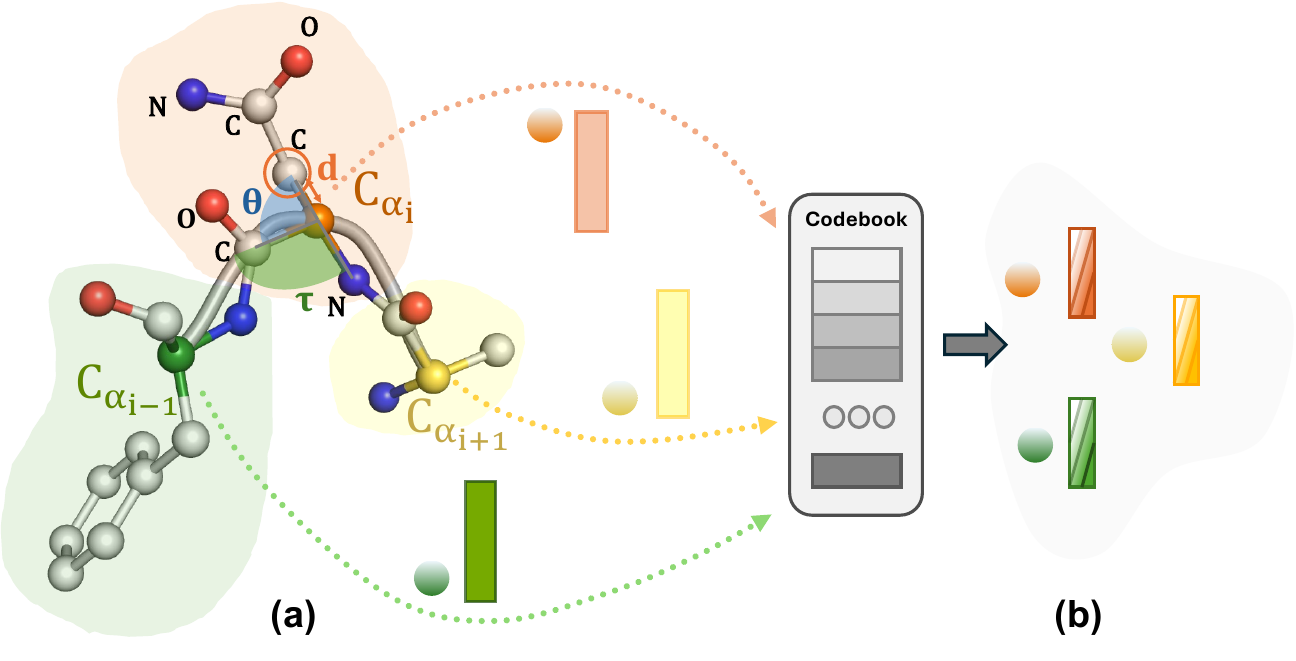}
    \caption{Illustration of the compression process. (a) Protein structures are compressed from all-atom representation to a low-dimensional graph with learned node features. (b) The continuous features are further discretized using a learned codebook for dimensionality reduction.}
    \label{fig:latent}
\end{figure}

Detailed pseudocode for training, inference, and internal-to-Cartesian reconstruction
is provided in the Supporting Information (Algorithms~S1--S4).

\subsection{Decoupling Constraints via Compression and Reconstruction}
\label{sec:compression}

Given a coarse-grained protein structure $\mathbf{CG} = \left\{ (X_i, A_i) \right\}_{i=1}^{N}$, protein backmapping reconstructs the full-atomic structure $\mathbf{x} \in \mathbb{R}^{n \times 3}$, where $n = \sum_{i=1}^N n_i$ denotes the total number of atoms. 
A key challenge lies in capturing diverse conformational states with high accuracy, while maintaining efficiency and generalization to unseen protein systems. 
Learning structural details and enforcing structural constraints directly in the all-atom space increases complexity and may degrade reconstruction quality, while post-generation refinement risks structural deviation. 
Moreover, generalizing to different conformational systems often leads to poor performance, as models struggle to capture the dynamic properties of unseen proteins.

Our approach focuses on obtaining effective protein structure representations that preserve structural validity and capture local variations through internal coordinates composed of bond lengths, bond angles, and torsion angles.
This strategy decouples structural constraints from the generative process, allowing them to be handled within a relatively simple and lightweight autoencoder stage.
To obtain the protein’s latent representation, we first compress the all-atom structure, a process illustrated in Figure~\ref{fig:latent}.

However, directly obtained protein representations exhibit low reconstruction accuracy. Unlike image representation where 2D adjacency sufficiently captures spatial relationships, protein structures require 3D geometric relationships between atoms, characterized by directional vectors $\mathbf{u}_{ij} = (\mathbf{x}_j - \mathbf{x}_i) / \|\mathbf{x}_j - \mathbf{x}_i\|$ and spatial distances $d_{ij} = \|\mathbf{x}_i - \mathbf{x}_j\|$. This implies that encoding merely global pairwise distances results in insufficient one-dimensional representations.

To better capture 3D geometric features during compression, we represent the all-atom structure $\mathbf{x} \in \mathbb{R}^{n \times 3}$ using a dual representation: a low-dimensional graph $\mathbf{X} \in \mathbb{R}^{N \times 3}$ and latent embeddings $\mathbf{h} \in \mathbb{R}^{N \times d}$. 
Here, $\mathbf{h}$ encodes relative atomic positions with respect to nodes in $\mathbf{X}$, while the graph preserves geometric relationships between residues. 
This dual representation reduces dimensionality while retaining essential 3D information. 
Additionally, structural constraints are naturally decoupled through geometric losses applied in both the encoder and decoder stages.

We realize this dual representation using an SE(3)-equivariant graph neural network with hierarchical message-passing~\citep{corso2022diffdock,yang2023chemically}. 
The architecture mirrors our two-level design by jointly operating on atomic and $C_\alpha$-level graphs to capture both local and global interactions.

\paragraph{Dual-level molecular graphs.}
We represent the protein as two coupled graphs 
$\mathcal{G}_{\text{atom}}=(V_a, E_a)$ and $\mathcal{G}_{\text{res}}=(V_r, E_r)$, 
processed jointly by an SE(3)-equivariant hierarchical GNN. 
At the \emph{atom level}, each node $i \in V_a$ represents an atom of type $a_i$ with initial embedding 
$\mathbf{h}_i^{(0)} = \mathrm{Embed}_a(a_i)$. 
Edges $(i,j)\in E_a$ connect atoms within a local cutoff radius $r_c^{\text{atom}} = 9\text{\AA}$, 
capturing bonded and near-neighbor interactions. 
Each edge feature encodes both radial and angular geometry:
\begin{equation}
\mathbf{e}_{ij}^{(0)} = 
\phi_d\!\left(\|\mathbf{x}_i - \mathbf{x}_j\|\right)
\oplus
\mathbf{Y}_l\!\left(\frac{\mathbf{x}_j - \mathbf{x}_i}{\|\mathbf{x}_j - \mathbf{x}_i\|}\right),
\end{equation}
where $\phi_d$ is a Gaussian distance expansion and $\mathbf{Y}_l$ are spherical-harmonic direction embeddings ensuring SE(3)-equivariance.

At the residue level, each node $k \in V_r$ represents a residue (C$_\alpha$ atom) 
with residue-type index $A_k$ and initial embedding 
$\mathbf{h}_k^{\alpha(0)} = \mathrm{Embed}_{r}(A_k)$. 
Residue edges $(k,l)\in E_r$ connect residues within a cutoff radius 
$r_c^{\text{res}} = 21\text{\AA}$ and are annotated by geometric features:
\begin{equation}
\mathbf{e}_{kl}^{(0)} = 
\phi_d\!\left(\big\|\mathbf{X}_k - \mathbf{X}_l\big\|\right)
\oplus
\mathbf{Y}_l\!\left(\frac{\mathbf{X}_l - \mathbf{X}_k}{\big\|\mathbf{X}_l - \mathbf{X}_k\big\|}\right).
\end{equation}
Cross-level edges $(i,k)\in E_{a\leftrightarrow r}$ connect each atom $i$ to its parent residue $A$ via the mapping $i\mapsto k$ and cutoff $r_c^{\text{cross}} = 21\text{\AA}$:
\begin{equation}
\mathbf{e}_{ik}^{\text{cross}(0)} =
\phi_c\!\left(\big\|\mathbf{x}_i - \mathbf{X}_k\big\|\right)
\oplus
\mathbf{Y}_l\!\left(\frac{\mathbf{X}_k - \mathbf{x}_i}{\big\|\mathbf{X}_k - \mathbf{x}_i\big\|}\right).
\end{equation}
These initialized node and edge features across all three connectivity types—atom, residue, and cross-level—serve as inputs to the subsequent SE(3)-equivariant hierarchical message passing network.

\paragraph{Hierarchical message passing.}
Each encoder layer alternates between intra-level and inter-level (atom–residue) message passing while maintaining SE(3)-equivariance. For atomic and residue features $\mathbf{h}_i^{(l)}$ and $\mathbf{h}_k^{\alpha(l)}$, the updates are defined as:
\begin{equation}
\begin{aligned}
\mathbf{h}_i^{(l+1)} &=
\mathbf{h}_i^{(l)} +
\phi_{\mathrm{atom}}\!\left(
\sum_{j \in \mathcal{N}_i^{a}} f_{\mathrm{atom}}(\mathbf{h}_j^{(l)}, \mathbf{e}_{ij})
\right)
+
\phi_{\mathrm{res}\rightarrow\mathrm{atom}}\!\left(
f_{\mathrm{cross}}(\mathbf{h}_{\alpha(i)}^{\alpha(l)}, \mathbf{e}_{i,\alpha(i)})
\right),\\
\mathbf{h}_k^{\alpha(l+1)} &=
\mathbf{h}_k^{\alpha(l)} +
\phi_{\mathrm{res}}\!\left(
\sum_{k' \in \mathcal{N}_k^{r}} f_{\mathrm{res}}(\mathbf{h}_{k'}^{\alpha(l)}, \mathbf{e}_{kk'})
\right)
+
\phi_{\mathrm{atom}\rightarrow\mathrm{res}}\!\left(
\sum_{j \in \mathcal{N}_k^{a}} f_{\mathrm{cross}}(\mathbf{h}_j^{(l)}, \mathbf{e}_{jk})
\right).
\end{aligned}
\label{eq:hierarchical_mp}
\end{equation}
Here, $\mathcal{N}_i^{a}$ and $\mathcal{N}_k^{r}$ denote atom- and residue-level neighborhoods, respectively, and $\mathcal{N}_k^{a}$ is the set of atoms belonging to residue $k$.  
The inter-level modules $\phi_{\mathrm{res}\rightarrow\mathrm{atom}}$ and $\phi_{\mathrm{atom}\rightarrow\mathrm{res}}$ enable bidirectional information exchange:  
residue-to-atom broadcasting and atom-to-residue aggregation.  
Each $\phi(\cdot)$ is implemented as a tensor-product convolution (\texttt{TensorProductConvLayer}), ensuring rotation- and translation-equivariance under $E(3)$ symmetry.  
After $L$ layers, atom and residue embeddings are fused and pooled per residue to produce $\mathbf{h}\in\mathbb{R}^{N\times d}$, the latent representation used for diffusion modeling.

\paragraph{Vector quantization.}
With the latent representations obtained, we further discretize them to enable efficient and stable generation. 
Specifically, we employ vector quantization~\citep{van2017neural} to map continuous features into discrete codes, which helps mitigate mode collapse observed in continuous VAEs~\citep{yang2023chemically}. 
A linear projection first reduces the feature dimension: $\mathbf{h}^\prime = \text{Linear}(\mathbf{h}) \in \mathbb{R}^{d^\prime}$ ($d^\prime < d$), improving computational efficiency and smoothing the latent space. 
The discrete representation $\mathbf{h}_q$ is then computed as:
\begin{equation}
\mathbf{h}_q = \arg\min_{\mathbf{e}_i \in \mathcal{E}} \|\mathbf{h}^\prime - \mathbf{e}_i\|^2
\end{equation}
where $\mathcal{E} = \{\mathbf{e}_1, \dots, \mathbf{e}_K\}$ is a learnable codebook of size $K$. 
This discrete encoding reduces generation complexity while preserving essential protein conformation features.

%%%%%%%%%%%%%%%%%%%%%%%%%%%%%%%%%%%%%%%%%%%%%%%%
%%%%%%%%%%%%%%%%%%%%%%%%%%%%%%%%%%%%%%%%%%%%%%%%

In the decoding phase, the decoder employs an SE(3)-invariant message passing network to map the residue-level latent representation $\mathbf{h}$ into per-residue internal geometric constraints 
$\mathcal{T} = \{\ell_{ij}, \theta_{ijk}, \phi_{ijkl}\}$~\citep{jing2022torsional}.
By operating on invariant pairwise relations such as inter-residue distances, the decoder captures geometry without being affected by global rotations or translations, while residue-type embeddings provide amino-acid–specific priors for local structure.
The predicted internal coordinates are then deterministically transformed into Cartesian coordinates, ensuring that reconstruction adheres to valid molecular geometry.
To ensure geometric fidelity and chemical plausibility, the training objective includes the following loss terms.

\textbf{Reconstruction losses} include bond length loss $\frac{1}{|B|} \sum_{b \in B} (b - \hat{b})^2$, bond angle loss $\frac{1}{|A|} \sum_{\theta \in A} \sqrt{2(1 - \cos(\theta - \hat{\theta})) + \epsilon}$, and torsion angle loss $\frac{1}{|T|} \sum_{\tau \in T} \sqrt{2(1 - \cos(\tau - \hat{\tau})) + \epsilon}$, where $\epsilon = 10^{-7}$ stabilizes gradients near extrema.

\textbf{Cartesian loss} enforces accuracy in atomic coordinates by computing the Root Mean Square Distance (RMSD) between predicted and ground-truth atom positions:
\begin{equation}
\mathcal{L}_{\text{xyz}} = \frac{1}{|N|} \sum_{\mathbf{x} \in N} \| \mathbf{x} - \hat{\mathbf{x}} \|_2^2
\end{equation}

\textbf{Clash loss} penalizes steric clashes between nonbonded atoms within a local 5 Å neighborhood:
\begin{equation}
\mathcal{L}_{\text{clash}} = 
\sum_{\mathbf{x}} \sum_{\mathbf{y} \in \mathcal{B}_r(\mathbf{x})} 
\max(2.0 - \| \mathbf{x} - \mathbf{y} \|_2^2, 0.0)
\end{equation}

\textbf{Graph loss} maintains local bonding geometry by minimizing changes in bond pair distances:
\begin{equation}
\mathcal{L}_{\text{graph}} = 
\frac{1}{|E|} \sum_{(i,j) \in E} 
\left( \| \mathbf{x}_i - \mathbf{x}_j \| - 
\| \hat{\mathbf{x}}_i - \hat{\mathbf{x}}_j \| \right)^2
\end{equation}

The total loss is:
\begin{equation}
\mathcal{L} = 
\mathcal{L}_{\text{recon}} +
\mathcal{L}_{\text{xyz}} +
\lambda_1 \mathcal{L}_{\text{clash}} +
\lambda_2 \mathcal{L}_{\text{graph}} +
\lambda_3 \mathcal{L}_{\text{vq}}
\label{eq:vaeloss}
\end{equation}
where $\mathcal{L}_{\text{recon}}$ refers to the sum of the three terms above, and $\mathcal{L}_{\text{vq}}$ is the vector quantization commitment loss~\citep{van2017neural}. We set $\lambda_1=5$, $\lambda_2=3$, and $\lambda_3=1$; note that $\mathcal{L}_{\text{clash}}$ and $\mathcal{L}_{\text{graph}}$ quickly converge to small values during training, making their scaling coefficients less sensitive.

This objective trains a constraint-consistent autoencoder, where the decoder $D$
deterministically maps latent variables to geometrically valid atomic structures,
thereby shaping a compact and physically meaningful latent manifold.
While this stage ensures geometric validity and structural consistency, it does not
directly model the conditional distribution $p(\mathbf{x}\mid\mathbf{X},\mathbf{A})$.
The conditional law is captured in the subsequent latent diffusion stage
(Eq.~\ref{eq:final_denoise_loss}), which learns
$p_\theta(\mathbf{h}\mid\mathbf{X},\mathbf{A})$; composing it with the deterministic
decoder $D$ then yields $p_\theta(\mathbf{x}\mid\mathbf{X},\mathbf{A})$ while
preserving geometric validity.

\subsection{Latent Diffusion in Compressed Space}
\label{sec:generation}

Traditional diffusion models for conformation ensemble often require complex constraint-handling (e.g., bond length enforcement or residue-wise denoising) to ensure geometric validity during generation. To simplify this, we perform diffusion in a learned latent space that inherently encodes structural priors, ensuring that the generative process follows the conditional distribution $p(\mathbf{x}\mid\mathbf{X},\mathbf{A})$.

Specifically, a VAE encoder $E$ maps all-atom structures $\mathbf{x}$ to latent features $\mathbf{h}_0 = E(\mathbf{x}) \in \mathcal{H}_{\text{struct}}$, a manifold where structural validity is preserved. 
Diffusion then proceeds directly in this latent space. 
During training, the forward process $q(\mathbf{h}_t\!\mid\!\mathbf{h}_{t-1})$ incrementally adds Gaussian noise, while the reverse process $p_\theta(\mathbf{h}_{t-1}\!\mid\!\mathbf{h}_t,\mathbf{X},\mathbf{A})$ learns to denoise conditioned on the CG graph.
Because the latent manifold $\mathcal{H}_{\text{struct}}$ already encodes geometric validity through Stage~(a), the reverse process does not require additional coordinate-level constraints, ensuring that generated structures remain physically consistent with the input CG configuration.

Note that Eq.~(\ref{eq:vaeloss}) defines the autoencoder reconstruction objective, ensuring that
the latent manifold preserves the geometric validity of $\mathbf{x}$.
The conditional distribution $p(\mathbf{x}\,|\,\mathbf{X},\mathbf{A})$ is learned
in the diffusion stage (Eq.~\ref{eq:final_denoise_loss}) through a denoising diffusion process within this latent space.

The model is trained with a standard noise prediction objective:
\begin{equation}
 \mathcal{L}_{\text{denoise}} = \mathbb{E}_{\mathbf{h}_0,\boldsymbol{\epsilon}\sim\mathcal{N}(0,1),t}\left[||\boldsymbol{\epsilon}-\epsilon_\theta(\mathbf{h}_t,t,\mathbf{c})||^2\right]
 \label{eq:final_denoise_loss}
\end{equation}
where the conditioning input $\mathbf{c}$ is the CG structure.

% % % % % % % % % % % % % % % % % % % % % % % % % % % % % % % % % % % % 

The conditional distribution $p(\mathbf{x}\mid\mathbf{X},\mathbf{A})$ is thus realized by composing the latent generative process with the deterministic decoder. 
Specifically, the autoencoder objective (Eq.~\ref{eq:vaeloss}) ensures that the decoder $D$ acts as a deterministic mapping approximately satisfying $p(\mathbf{x}\mid \mathbf{h},\mathbf{X},\mathbf{A}) \approx \delta(\mathbf{x}-D(\mathbf{h};\mathbf{X},\mathbf{A}))$. 
Meanwhile, the diffusion objective (Eq.~\ref{eq:final_denoise_loss}) learns the conditional latent distribution $p_\theta(\mathbf{h}\mid\mathbf{X},\mathbf{A})$. 
By integrating over the latent space, the generation process samples from the target distribution:
\begin{equation}
p(\mathbf{x}\mid\mathbf{X},\mathbf{A})
\;=\;
\int
p(\mathbf{x}\mid \mathbf{h},\mathbf{X},\mathbf{A})
\,p_\theta(\mathbf{h}\mid\mathbf{X},\mathbf{A})
\,d\mathbf{h}
\;\approx\;
\int \delta\!\big(\mathbf{x}-D(\mathbf{h};\mathbf{X},\mathbf{A})\big)
\,p_\theta(\mathbf{h}\mid\mathbf{X},\mathbf{A})
\,d\mathbf{h}.
\end{equation}
Therefore, sampling $\mathbf{h}$ via the learned reverse diffusion and decoding it yields structures that follow the desired conditional distribution while strictly adhering to the geometric constraints imposed by the decoder.

% % % % % % % % % % % % % % % % % % % % % % % % % % % % % % % % % % % % 

Our denoising network $\epsilon_\theta$, based on ProteinMPNN~\citep{dauparas2022robust}, takes four inputs: the noisy latent representation $\mathbf{h}_t$, CG coordinates $\mathbf{X}$, protein sequence $\mathbf{A}$, and diffusion timestep $t$.
To guide consistent denoising of $\mathbf{h}_t$, $\mathbf{X}$ and $\mathbf{A}$ serve as node and edge features within the graph architecture, while $t$ is processed via adaptive layer normalization (adaLN)~\citep{perez2018film} for time-dependent modulation in normalization layers.

Our process directly updates latent features $\mathbf{h}_t$, not indices, allowing dynamic adjustments of structural elements (e.g., sidechains responding to inter-atomic interactions) during refinement.
Furthermore, it operates non-autoregressively, updating all latent features in parallel conditioned on the global CG structure.

Operating in this VAE-structured latent space enables efficient generation and strong generalization across diverse conformations, as the diffusion process samples abstract latent features while the manifold inherently ensures geometric precision.

\section{Results and Discussion}

\subsection{Experimental Setup}

\paragraph{Datasets.} 
(1) \textbf{PED}~\citep{lazar2021ped,ghafouri2024ped} contains structural ensembles of intrinsically disordered proteins (IDPs). Following~\citep{yang2023chemically}, this study selected 85 proteins with approximately 100 conformations each. The test set includes four proteins consistent with previous work: PED00151ecut0, PED00090e000, PED00055e000, and PED00218e000. 
(2) \textbf{ATLAS}~\citep{vander2024atlas} provides MD simulations across 1,390 non-membrane proteins, covering all ECOD structural classes~\citep{schaeffer2017ecod}. Each protein includes three 100-ns replicate simulations, with 300 randomly sampled conformations. After filtering sequences longer than 512 residues, 1,297 proteins remained for training, with 70 proteins (PDB entries after May 1, 2019) held out for testing. 
(3) \textbf{PDB}~\citep{berman2000protein} dataset, extended through SidechainNet~\citep{king2021sidechainnet}, contains single-conformation protein structures. Following~\citep{jones2023diamondback}, the dataset was filtered to removed incomplete structures, inconsistent entries, and long sequences. The final dataset contains 62,105 proteins for training and 23 for testing. 
(4) \textbf{DES}~\citep{lindorff2011fast} provides all-atom MD trajectories for fast-folding proteins. Our test set uses 11 of these proteins (excluding 2F4K due to its non-canonical amino acid, following~\citep{jones2023diamondback,jones24flowback}), sampling 10,000 equally spaced frames from one trajectory for each. This dataset evaluates model performance on challenging, dynamic structures that represent real-world conformational diversity.

\paragraph{Baselines.} The paper compares against three recent backmapping methods: 
GenZProt~\citep{yang2023chemically} aligns all-atom and coarse-grained structures in a shared latent space using dual VAE encoders, but its Gaussian prior limits conformational diversity. 
DiAMoNDBack~\citep{jones2023diamondback} uses an autoregressive diffusion framework. This approach achieves effective reconstruction but leads to error accumulation and computational overhead. 
FlowBack~\citep{jones24flowback} uses a flow matching training objective for denoising directly in the all-atom space, yet it shows limited generalizability across multi-conformational spaces.

\paragraph{Metrics.} 
The model is evaluated using common metric as backmapping method~\citep{yang2023chemically,jones2023diamondback,jones24flowback}. 
(1) \textbf{RMSD} evaluates the reconstruction quality by measuring the average distance between corresponding atoms in generated and
reference structures.
(2) \textbf{Graph Edit Distance (GED)} measures the preservation of chemical bond topology by computing distance differences between bonded atoms.
(3) \textbf{Steric Clash Score} quantifies the physical plausibility by mea-
suring the ratio of atomic overlaps: where a clash is defined as an atom-atom distance smaller than 1.2 Å.
(4) \textbf{Diversity Score (DIV)}: As backmapping is inherently a one-to-many operation, an ideal model should generate a diverse ensemble of physically plausible structures. The DIV score measures structural variation among generated samples:
\begin{equation}
\begin{aligned}
\text{RMSD}_\text{ref} &= \frac{1}{G} \sum_{i=1}^{G} \text{RMSD}(\mathbf{x}_{i}^\text{gen}, \mathbf{x}_{i}^\text{ref}) \\
\text{RMSD}_\text{gen} &= \frac{2}{G(G-1)} \sum_{i=1}^{G} \sum_{j<i} \text{RMSD}(\mathbf{x}_{i}^\text{gen}, \mathbf{x}_{j}^\text{gen}) \\
\text{DIV} &= 1 - \frac{\text{RMSD}_\text{gen}}{\text{RMSD}_\text{ref}}
\end{aligned}
\end{equation}
where $G$ is the number of generated structures.

Additionally, (5) \textbf{Computational Efficiency} is evaluated by measuring the total inference time required to perform backmapping ten times for each structure in the test sets. Lower values indicate better performance for all metrics.

\paragraph{Implementation Details.} For CODLAD's autoencoder, we adopted parameters from~\citep{yang2023chemically}, setting the output node feature dimension to 36. The vector quantization stage utilized a 4096-entry codebook with an embedding dimension of 3. This autoencoder was trained with a batch size of 4, using initial learning rates of 0.001 for the PED and PDB datasets, and 0.0005 for the ATLAS dataset.

The diffusion framework employed a linear variance schedule ($t_{\text{max}} = 1000$, variance ranging from $1 \times 10^{-4}$ to $2 \times 10^{-2}$) and a learned covariance $\Sigma_{\theta}$, following~\citep{peebles2023scalable}. During sampling, 100 steps were used to balance computational efficiency with output quality. The denoising neural network featured a 3-layer encoder-decoder architecture with a hidden layer size of 128. This network was trained using a learning rate of $3 \times 10^{-4}$, a batch size of 128, a 20,000-step warmup period, and a linear learning rate schedule for 300,000 steps, eventually annealing to $1 \times 10^{-5}$. All models were trained and evaluated on a single NVIDIA A100 GPU with 40GB of memory.

% ped55
%%
\begin{figure*}[t]
    \centering
    \includegraphics[width=\textwidth]{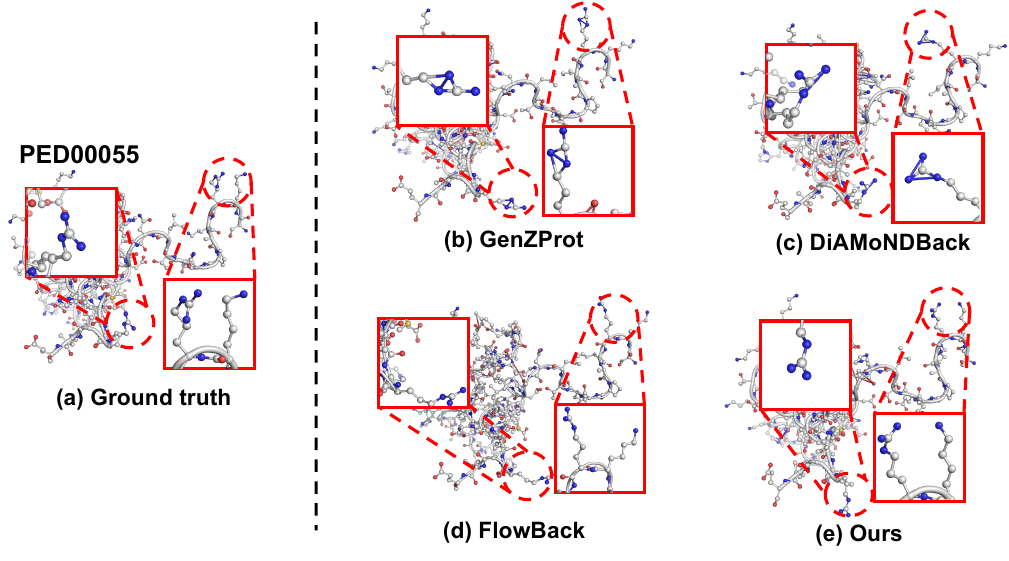} 
    \caption{Visualization of PED00055 protein structure generation from the PED dataset. Our method (e) maintains accurate structural validity near flexible side chains (red circles), closely matching the ground truth (a). In contrast, GenZProt (b) and DiAMoNDBack (c) generate conflicting side chain atoms in these regions, while FlowBack (d) does not produce obvious conflicts but yields erroneous backbone topology (red C$_\alpha$ atoms), which may result from directly performing diffusion in the full-atom space.}

    \label{fig:ped55}
\end{figure*}

\subsection{Constraint Decoupling Improves Accuracy and Diversity}
\label{sec:mainresults}

Decoupling structural constraints from the generative step allows the diffusion model to explore a broad conformational space and reach atomic-level accuracy while maintaining structural and topological consistency. 
Table~\ref{table:main} summarizes results on multiple protein datasets, where CODLAD yields consistent improvements in reconstruction and diversity metrics.

CODLAD reduces RMSD by 4.3\%--10.4\%, reflecting the iterative denoising behavior of diffusion. Because backmapping recover subtle atomistic details from coarse-grained inputs, RMSD typically spans a narrow range; at an \emph{atomic-level} precision, an improvement of this magnitude is meaningful and indicates reliable recovery of full-atom structure.

\begin{table}[th]
\caption{Quantitative comparison of different backmapping methods across multiple datasets. Lower values indicate better performance for all metrics, while DIV should be interpreted alongside reconstruction quality. Best results are shown in bold. CODLAD achieves the lowest RMSD and substantially improves topological consistency, with over 40\% GED reduction on dynamic datasets. It remains competitive on steric plausibility, where low clash from other methods reflects expanded coordinates that reduce steric contacts but do not improve RMSD or GED. On ATLAS, CODLAD also lowers DIV, showing that constraint decoupling improves both accuracy and diversity without compromising structural fidelity.}
\label{table:main}
\begin{center}\small
\begin{small}
\begin{sc}
    \begin{tabular}{lccccccc}
    \toprule
    & Method & PED & PDB & ATLAS \\
    \midrule
    \multirow{3}{*}{RMSD ($\downarrow$)}
    & GenZProt & 
    1.834\scriptsize{$\pm$0.157} & 
    1.610\scriptsize{$\pm$0.162} &
    1.718\scriptsize{$\pm$0.157} \\
    & DiAMoNDBack & 
    1.802\scriptsize{$\pm$0.117} &
    1.294\scriptsize{$\pm$0.192} &
    3.165\scriptsize{$\pm$0.086} \\
    & FlowBack & 
    1.787\scriptsize{$\pm$0.130} &
    1.366\scriptsize{$\pm$0.169} &
    3.240\scriptsize{$\pm$0.091} \\
    & Ours & 
    \textbf{1.710\scriptsize{$\pm$0.089}} &
    \textbf{1.236\scriptsize{$\pm$0.183}} &
    \textbf{1.539\scriptsize{$\pm$0.176}} \\
    
    \midrule
    \multirow{3}{*}{GED ($\downarrow$)}
    & GenZProt & 
    0.800\scriptsize{$\pm$0.022} &
    0.393\scriptsize{$\pm$0.058} &
    0.715\scriptsize{$\pm$0.188} \\
    & DiAMoNDBack & 
    4.385\scriptsize{$\pm$0.032} &
    0.714\scriptsize{$\pm$0.003} &
    3.653\scriptsize{$\pm$0.291} \\
    & FlowBack & 
    4.865\scriptsize{$\pm$0.199} &
    0.171\scriptsize{$\pm$0.102} &
    3.913\scriptsize{$\pm$0.325} \\
    & Ours & 
    \textbf{0.471\scriptsize{$\pm$0.077}} &
    \textbf{0.162\scriptsize{$\pm$0.093}} &
    \textbf{0.391\scriptsize{$\pm$0.044}} \\
    
    \midrule
    \multirow{3}{*}{Clash ($\downarrow$)}
    & GenZProt & 
    0.448\scriptsize{$\pm$0.044} &
    0.660\scriptsize{$\pm$1.123} &
    0.232\scriptsize{$\pm$0.265} \\
    & DiAMoNDBack & 
    1.087\scriptsize{$\pm$0.009} &
    0.422\scriptsize{$\pm$0.905} &
    \textbf{0.018\scriptsize{$\pm$0.010}} \\
    & FlowBack & 
    0.353\scriptsize{$\pm$0.043} &
    \textbf{0.413\scriptsize{$\pm$0.902}} &
    0.022\scriptsize{$\pm$0.014} \\
    & Ours & 
    \textbf{0.325\scriptsize{$\pm$0.044}} &
    0.435\scriptsize{$\pm$0.907} &
    0.047\scriptsize{$\pm$0.096} \\

    \midrule
    \multirow{3 }{*}{DIV ($\downarrow$)}
    & GenZProt &
    0.898\scriptsize{$\pm$0.008} &
    0.468\scriptsize{$\pm$0.020} &
    0.632\scriptsize{$\pm$0.023} \\
    & DiAMoNDBack &
    0.484\scriptsize{$\pm$0.001} &
    \textbf{0.367\scriptsize{$\pm$0.031}} &
    0.701\scriptsize{$\pm$0.018} \\
    & FlowBack &
    0.472\scriptsize{$\pm$0.030} &
    0.487\scriptsize{$\pm$0.029} &
    0.771\scriptsize{$\pm$0.015} \\
    & Ours &
    \textbf{0.452\scriptsize{$\pm$0.020}} &
    0.379\scriptsize{$\pm$0.034} &
    \textbf{0.476\scriptsize{$\pm$0.029}} \\
    
    \bottomrule
    \end{tabular}
\end{sc}
\end{small}
\end{center}
\end{table}

Topological consistency benefits primarily from the smoother latent space shaped by constraint compression. On the large, dynamic ATLAS dataset, CODLAD surpasses baselines that denoise directly in atomic coordinates, which struggle to capture conformational variability. Notably, DiAMoNDBack and FlowBack underperform on GED—nearly losing topological consistency—although they remain stable on the static dataset, suggesting that coordinate-space denoising does not explicitly respect graph structure. By injecting structural constraints during latent space compression, CODLAD avoids this issue and achieves large improvements on dynamic datasets—41.1\% and 45.3\% reductions in GED on PED and ATLAS, respectively.

For steric plausibility, CODLAD is competitive across datasets, but gains are smaller than for accuracy and topology. On ATLAS, DiAMoNDBack and FlowBack report low clash despite poor RMSD and GED; this likely arises because all-atom diffusion started from CG structure expands coordinates and lowers packing density, yielding low clash scores irrespective of alignment quality. Clash chiefly reflects intra-structure sterics and is only weakly coupled to the target conformation, so improving clash alone does not guarantee accurate backmapping. In our setting, the latent space clash term contributes less than topology constraints, which drive most improvements; accordingly, CODLAD prioritizes geometric and topological fidelity while keeping low steric clashes.

\begin{table}[b!]
\caption{Generalization performance on the unseen DES fast-folding dataset after pretraining on PDB. CODLAD achieves the lowest RMSD and GED, reducing them by 12.6\% and 56.9\% relative to the best baseline, and also improves diversity while keeping steric clashes low. In contrast, DiAMoNDBack and FlowBack yield very low clash but extremely high GED, showing that expanded coordinates reduce steric contacts without preserving topology. These results indicate that constraint decoupling in latent space yields more reliable generalization across trajectory systems.}
\label{table:des} 
\begin{center}\small
\begin{small}
\begin{sc}
    \begin{tabular}{lcccc}
    \toprule
    Method & RMSD ($\downarrow$) & GED ($\downarrow$) & Clash ($\downarrow$) & DIV ($\downarrow$) \\ 
    \midrule
    GenZProt & 
    1.959\scriptsize{$\pm$0.260} & 
    1.228\scriptsize{$\pm$0.990} & 
    1.102\scriptsize{$\pm$0.894} & 
    0.410\scriptsize{$\pm$0.016} \\
    DiAMoNDBack & 
    2.115\scriptsize{$\pm$0.763} & 
    11.049\scriptsize{$\pm$1.700} & 
    0.367\scriptsize{$\pm$0.562} & 
    0.469\scriptsize{$\pm$0.106} \\
    FlowBack & 
    2.140\scriptsize{$\pm$0.801} & 
    12.578\scriptsize{$\pm$1.967} & 
    \textbf{0.005\scriptsize{$\pm$0.001}} & 
    0.592\scriptsize{$\pm$0.091} \\
    Ours & 
    \textbf{1.713\scriptsize{$\pm$0.232}} & 
    \textbf{0.529\scriptsize{$\pm$0.076}} & 
    0.049\scriptsize{$\pm$0.040} & 
    \textbf{0.381\scriptsize{$\pm$0.178}} \\
    \bottomrule
    \end{tabular}
\end{sc}
\end{small}
\end{center}
\end{table}

For diversity, CODLAD performs strongly on the large, dynamic ATLAS dataset, achieving a 24.7\% improvement in DIV and indicating that generation in a smoother latent space remains robust as conformational complexity increases. 
On the smaller dynamic set PED, all-atom diffusion methods attain lower DIV than the VAE baseline, but on ATLAS their DIV worsens, suggesting that multi-step inference mitigates mode collapse in simpler settings yet yields less varied structures as complexity grows. 
In contrast, CODLAD keeps diversity controlled across dynamic and static datasets without degrading accuracy or topology. Overall, this pattern indicates that latent space generation is well suited to practical scenarios with data sparsity or uneven coverage of the conformational landscape.

\subsection{Robust Generalization to Unseen Trajectory Systems}
\label{sec:generalization_exp}

Generalization stems from learning and sampling in a discrete latent space with structural constraints, which regularizes the manifold and improves transfer under distribution shift. 
We therefore evaluate all methods on an unseen fast-folding dataset (DES) after pretraining on PDB. Although PDB is primarily static, its broad structural coverage provides a suitable foundation for training general-purpose models. Table~\ref{table:des} summarizes the results.

On DES, CODLAD achieves the best reconstruction accuracy and topology. RMSD and GED are reduced by $12.6\%$ and $56.9\%$ relative to the strongest baseline, and DIV decreases by $7.1\%$. Methods that denoise directly in all-atom coordinates (DiAMoNDBack, FlowBack) show high GED, indicating weak preservation of graph connectivity. Their low clash values likely result from diffusion initialized from coarse-grained structures, which increases interatomic spacing and lowers packing density; low clash does not imply accurate backmapping. Decoupling constraints and generating in latent space improves generalization across trajectory systems and is useful when simulation protocols differ or data coverage is sparse.

\subsection{Latent Constraints Match Torsion Distributions}
\label{sec:dis}

\begin{figure}[thbp]
    \centering
    \includegraphics[width=\columnwidth]{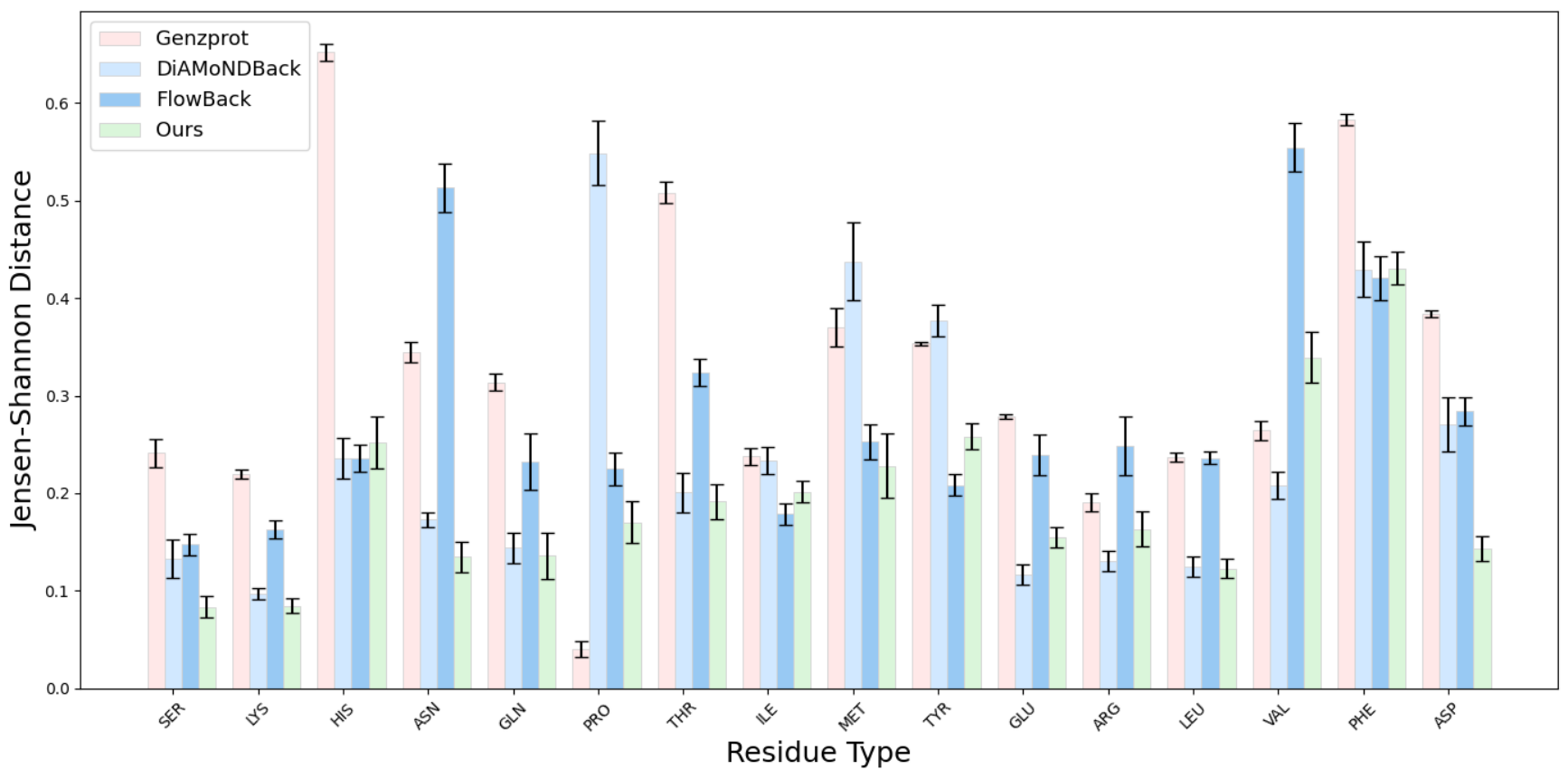}
    \caption{Residue-Wise Divergence for Protein PED00055. JSD of torsion angle distributions for each residue in protein PED00055. CODLAD achieves consistently lower JSD across residue types, particularly on flexible side chains (e.g., ASP, GLU, PHE), and avoids extreme outliers observed in baseline methods, highlighting more reliable residue-level geometry modeling.}
    \label{fig:jsd}
\end{figure}
\begin{figure}[thbp]
    \centering
    \includegraphics[width=0.7\columnwidth]{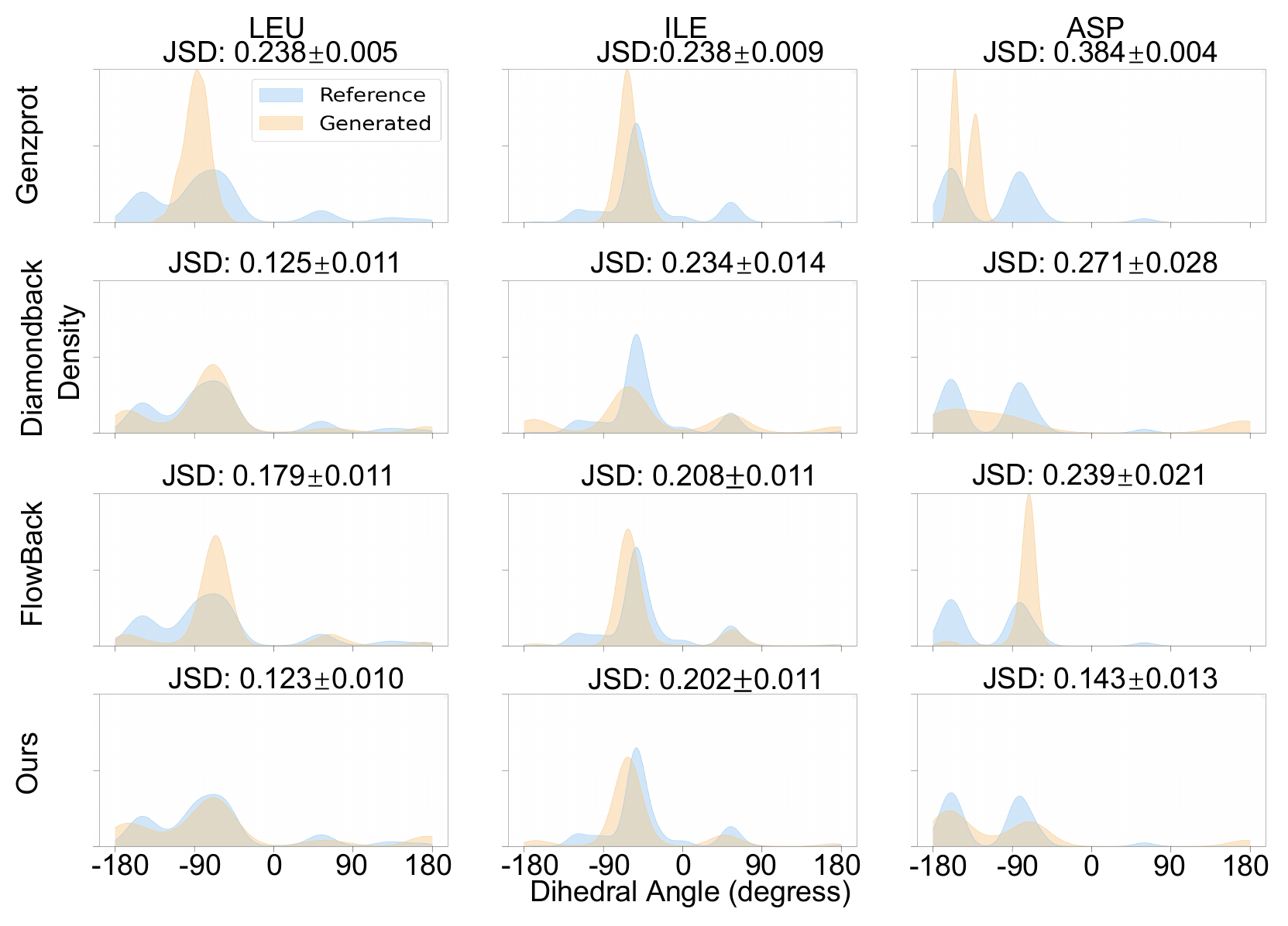}
    \caption{Torsion Angle Distributions for Selected Residues. KDE of torsion angle distributions for LEU, ILE, and ASP residues. CODLAD better aligns with the ground truth and accurately recovers multimodal rotamer patterns. In contrast, baselines often blur or miss peaks, especially for residues with complex side-chain conformations such as ASP.}
    \label{fig:dis}
\end{figure}

Our method closely matches torsion-angle distributions and keeps bond/dihedral geometry tighter than competing methods because structural constraints are learned in the latent space: bond lengths, bond angles, and dihedrals are encoded during compression, so samples respect them even without all-atom denoising. On the PED00055 conformational ensemble—selected to enable direct, like-for-like comparison with prior work, we evaluate $\chi_1$ torsions (excluding residues without $\chi_1$, e.g., Gly/Ala), estimate predicted vs.\ reference densities with Kernel Density Estimation (KDE), quantify similarity via Jensen--Shannon divergence (JSD), and assess temporal stability using frame-wise bond and torsion angle deviations over the full trajectory.

\begin{figure*}[thb]
    \centering
    \includegraphics[width=1\textwidth]{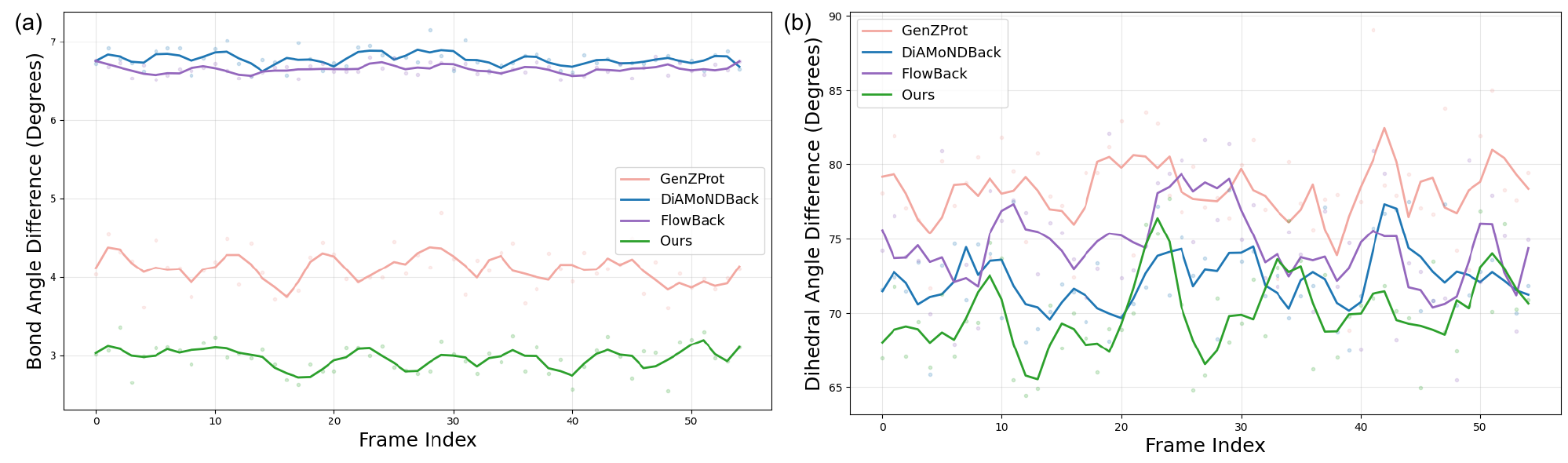}
    \caption{Comparison of bond angle and torsion angle deviations across different conformational states of PED00055. Lower values indicate better preservation of ideal bond geometry. CODLAD maintains consistently smaller deviations than baseline methods, with bond angles closely matching reference values and torsion angles exhibiting more stable trajectories over time, confirming improved geometric fidelity across the ensemble.}
    \label{fig:angle_torsion}
\end{figure*}

As shown in Figure~\ref{fig:jsd}, analysis of $\chi_1$ angle distributions in test protein PED00055 reveals consistently lower JSD values across residue types compared to baseline methods. 
The improved distribution matching is particularly evident in residues with complex side chains (ASP, GLU, and PHE), where accurate torsion angle modeling is essential for structural validity.

Figure~\ref{fig:dis} shows fitted $\chi_1$ distributions for representative residues. For LEU and ILE, our method recovers the expected multimodal rotameric patterns, whereas the baselines blur peaks or miss entire modes—consistent with averaging across periodic states. For ASP, whose side chain exhibits richer conformational preferences, CODLAD likewise preserves the distinct rotamer peaks rather than smoothing them away.

We further analyze the geometric accuracy across all frames of protein PED00055, which represents different conformations of the protein over time. As shown in Figure~\ref{fig:angle_torsion}, our method maintains lower bond angle deviations throughout the trajectory compared to both baseline methods. For dihedral angles, which are more flexible and challenging to model, our method demonstrates more stable behavior with consistently lower deviations. 
These results across multiple conformational states validate our framework's ability to maintain geometric accuracy while capturing natural conformational variations.

\subsection{Accelerated Inference with Latent Denoising}
\label{sec:time}

\begin{figure}[t!]
    \centering
    \includegraphics[width=0.6\columnwidth]{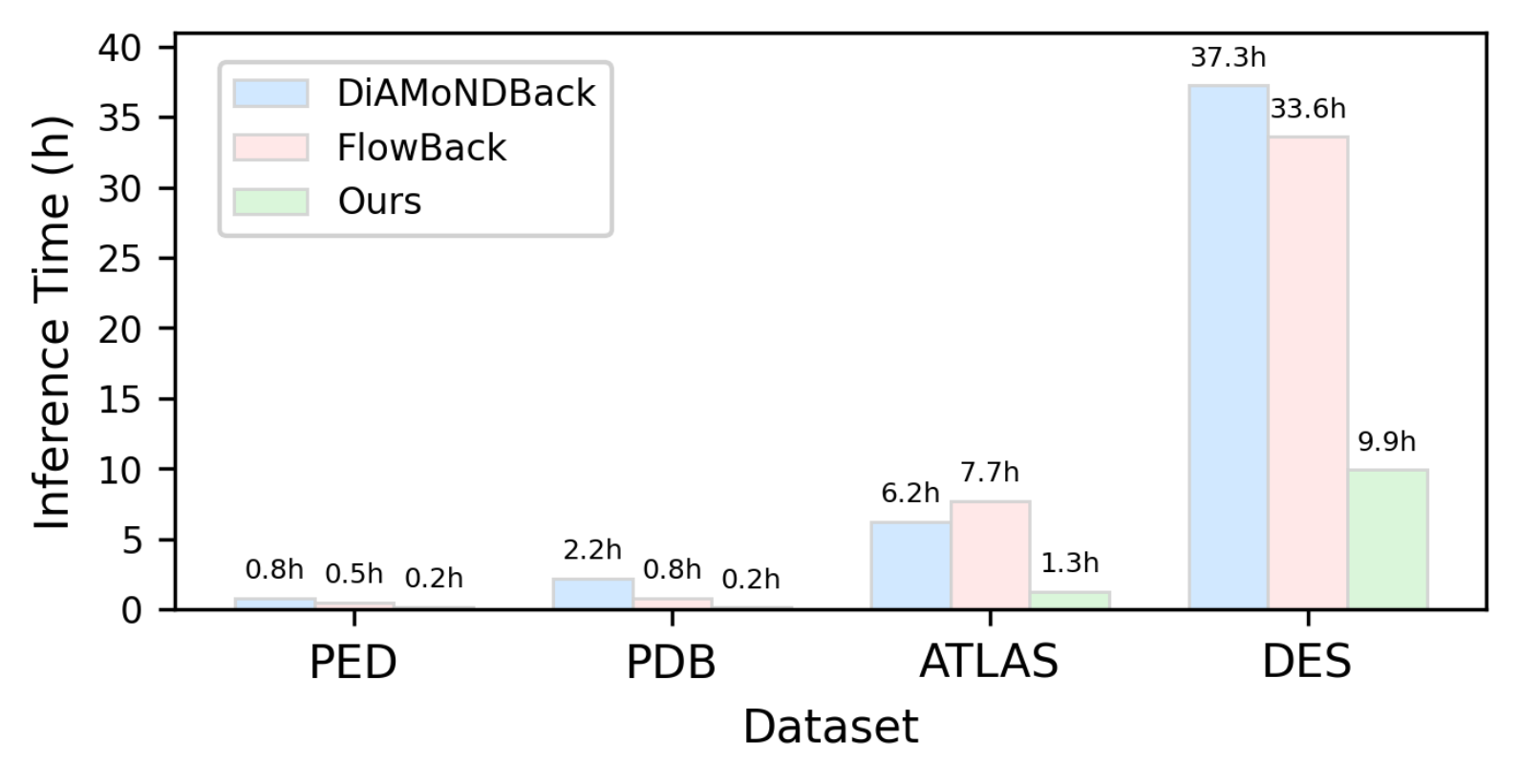}
    \caption{Comparison of inference time across different testsets. As a diffusion-based method, our approach achieves substantial efficiency improvements over the existing multi-step method.}
    \label{fig:inference_time}
\end{figure}

CODLAD achieves markedly lower inference time than all-atom diffusion methods. Total runtime is reduced by roughly 70\% across datasets (typically 60--80\%). The improvement results from denoising low dimensional latent representations rather than atomic coordinates, which reduces the number of nodes and edges processed and the cost of pairwise computations. In the inference evaluation, sampling is performed once on the test set, and ten samples are generated for each input. The corresponding runtimes are shown in Figure~\ref{fig:inference_time}.

As observed, for datasets such as ATLAS and DES, all-atom reverse mapping of the test set is prohibitively slow. Even for DES proteins (10--80 residues), all-atom denoising can take more than a day, even with trajectory subsampling, which limits usability and slows experimental progress. CODLAD markedly accelerates this workflow: on DES it reduces runtime by about $70\%$, saving approximately $24$ hours per test set. For pipelines that backmapping long simulation trajectories, the shorter turnaround increases throughput and makes routine processing feasible without sacrificing accuracy or topological consistency.

\begin{table}[th]
\caption{Ablation study on the PED dataset evaluating the effect of structural constraints during compression. The baseline “Basic” includes bond length, bond angle, and torsion constraints from internal coordinates. Adding GED and Clash constraints progressively improves reconstruction accuracy and structural validity: GED reduces topological errors, Clash minimizes steric clashes, and both combined yield the best trade-off across all metrics. The DIV metric should be interpreted jointly with accuracy and validity. Best results are in bold, second-best underlined.} 
\label{table:ablation_constrain} 
\begin{center}\small 
\begin{sc} 
    \begin{tabular}{lcccc}
    \toprule
    Constraints & RMSD ($\downarrow$) & GED ($\downarrow$) & Clash ($\downarrow$) & DIV ($\downarrow$) \\
    \midrule
    Basic & 
    1.781\scriptsize{$\pm$0.102} & 
    1.817\scriptsize{$\pm$1.043} & 
    0.404\scriptsize{$\pm$0.050} & 
    \textbf{0.373\scriptsize{$\pm$0.026}} \\
    + GED & 
    1.739\scriptsize{$\pm$0.114} & 
    \underline{0.477\scriptsize{$\pm$0.192}} & 
    0.378\scriptsize{$\pm$0.047} & 
    0.448\scriptsize{$\pm$0.026} \\
    + Clash & 
    1.717\scriptsize{$\pm$0.103} & 
    1.214\scriptsize{$\pm$0.595} & 
    \underline{0.375\scriptsize{$\pm$0.048}} & 
    0.422\scriptsize{$\pm$0.031} \\
    \rowcolor{gray!20} + Both & 
    \textbf{1.710\scriptsize{$\pm$0.089}} & 
    \textbf{0.471\scriptsize{$\pm$0.077}} & 
    \textbf{0.325\scriptsize{$\pm$0.044}} & 
    0.452\scriptsize{$\pm$0.020} \\
    \bottomrule
    \end{tabular}
\end{sc}
\end{center}
\end{table}

\subsection{Ablation Study}
\label{sec:ablation}

\subsubsection{Effectiveness of Structural Constraint Decoupling}

To evaluate structural constraint decoupling mechanism, we conducted ablation studies on the PED datasets. We progressively incorporated geometric (GED) and clash (Clash) constraints during the compression phase, with results detailed in Table~\ref{table:ablation_constrain}. The baseline 'basic' configuration inherently includes constraints on bond lengths, angles, and torsion angles, as our model utilizes internal coordinates.

\begin{table}[th] 
\caption{Ablation study of vocabulary size and hidden dimension on PED conformational ensembles. Results show that performance improves as vocabulary grows up to 4096 entries but degrades at 8192, indicating diminishing utilization with overly large codebooks. For hidden dimension, optimal accuracy and validity are achieved at size 3, with both smaller (under-capacity) and larger (overfitting/optimization difficulty) dimensions leading to worse results. Best results are bold.}

\label{table:ablation_hidden_vocab}
\begin{center}\small 
\begin{sc} 
    \begin{tabular}{lcccc} 
    \toprule
    Setting & RMSD ($\downarrow$) & GED ($\downarrow$) & Clash ($\downarrow$) & DIV ($\downarrow$) \\
    \midrule
    \multicolumn{5}{l}{\textbf{Vocabulary Size Ablation}} \\
    512 & 
    1.766\scriptsize{$\pm$0.130} & 
    1.111\scriptsize{$\pm$0.254} & 
    0.955\scriptsize{$\pm$0.049} & 
    0.393\scriptsize{$\pm$0.025} \\
    2048 & 
    1.888\scriptsize{$\pm$0.137} & 
    0.547\scriptsize{$\pm$0.120} & 
    0.394\scriptsize{$\pm$0.041} & 
    \textbf{0.350\scriptsize{$\pm$0.038}} \\
    \rowcolor{gray!20} 4096 & 
    \textbf{1.710\scriptsize{$\pm$0.089}} & 
    \textbf{0.471\scriptsize{$\pm$0.077}} & 
    \textbf{0.325\scriptsize{$\pm$0.044}} & 
    0.452\scriptsize{$\pm$0.020} \\
    8192 & 
    1.790\scriptsize{$\pm$0.129} & 
    0.607\scriptsize{$\pm$0.115} & 
    0.384\scriptsize{$\pm$0.043} & 
    0.376\scriptsize{$\pm$0.032} \\
    \multicolumn{5}{l}{\textbf{Hidden Size Ablation}} \\ 
    1 & 
    1.880\scriptsize{$\pm$0.146} & 
    0.824\scriptsize{$\pm$0.135} & 
    0.439\scriptsize{$\pm$0.042} & 
    0.398\scriptsize{$\pm$0.044} \\
    \rowcolor{gray!20} 3 & 
    \textbf{1.710\scriptsize{$\pm$0.089}} & 
    \textbf{0.471\scriptsize{$\pm$0.077}} & 
    \textbf{0.325\scriptsize{$\pm$0.044}} & 
    0.452\scriptsize{$\pm$0.020} \\
    4 & 
    1.748\scriptsize{$\pm$0.124} & 
    0.590\scriptsize{$\pm$0.128} & 
    0.349\scriptsize{$\pm$0.044} & 
    0.410\scriptsize{$\pm$0.035} \\
    9 & 
    1.776\scriptsize{$\pm$0.109} & 
    0.631\scriptsize{$\pm$0.113} & 
    0.331\scriptsize{$\pm$0.045} & 
    0.439\scriptsize{$\pm$0.033} \\
    18 & 
    1.734\scriptsize{$\pm$0.096} & 
    0.597\scriptsize{$\pm$0.049} & 
    0.341\scriptsize{$\pm$0.043} & 
    0.390\scriptsize{$\pm$0.028} \\
    36 & 
    1.808\scriptsize{$\pm$0.142} & 
    0.565\scriptsize{$\pm$0.034} & 
    0.421\scriptsize{$\pm$0.041} & 
    \textbf{0.368\scriptsize{$\pm$0.037}} \\
    \bottomrule
    \end{tabular}
\end{sc}
\end{center}
\end{table}

The ablation results in Table~\ref{table:ablation_constrain} highlight the contribution of structural constraint decoupling. Starting from the “Basic” configuration, which already respects local internal coordinates, adding GED constraints dramatically reduces topological errors and simultaneously lowers clashes due to improved connectivity. Incorporating Clash constraints further suppresses residual steric violations. With both constraints, RMSD is minimized and overall validity improved, confirming that constraints are effectively captured in the compressed representation rather than enforced post hoc. Although diversity (DIV) decreases slightly, the trade-off favors closer structural alignment with targets while maintaining sufficient conformational variability. These findings demonstrate that learning constraints during compression provides a principled way to balance accuracy, validity, and diversity without hand-tuned penalties in all-atom space.

In summary, constraint decoupling in latent space achieves the best balance between accuracy, validity, and diversity, highlighting its value as a principled design choice.

\subsubsection{Balancing Vocabulary Size and Hidden Dimension}

We conducted ablation studies on PED conformational ensembles to assess how vocabulary size and hidden dimension of discrete representations impact model performance, offering insights for parameter optimization.

Our analysis of vocabulary size (Table~\ref{table:ablation_hidden_vocab}) indicates that an optimal size consistently emerges around 4096 across various metrics and models. Increasing the vocabulary size from 512 to 4096 yields substantial performance improvements. However, further expansion, for instance to 8192, results in performance degradation. This decline suggests challenges in efficient vocabulary utilization. Specifically, while the reconstruction accuracy (e.g., of a VQ-VAE) may continue to improve with larger vocabularies, the quality of generated samples tends to decrease. This discrepancy might stem from an inconsistency between optimizing for reconstruction versus generation, where excessively large vocabularies lead to low utilization rates and hinder effective learning of the latent representations.

Regarding hidden dimensions (Table~\ref{table:ablation_hidden_vocab}), optimal performance is consistently achieved with a dimension of approximately 3. Performance deteriorates at both extremes: a hidden dimension of 1 results in significantly higher error rates across all metrics, indicating insufficient representational capacity, whereas larger dimensions (e.g., 18 or 36) lead to a gradual decline in performance, potentially due to overfitting or increased optimization challenges.

Together these findings indicate that vocabulary size and hidden dimension act in tandem: too small a capacity underfits, while too large wastes representation power, and the optimal balance lies in a compact yet expressive latent space with high code utilization.

\subsubsection{Diffusion and Discretization Improve Performance}

\begin{table}[t] 
\caption{Ablation on PED dataset for model architecture. Results compare continuous vs.\ discrete latent spaces (VAE vs.\ VQ-VAE) and generative dynamics (diffusion vs.\ flow). VQ-VAE with diffusion achieves the best balance across all metrics, showing that discretization improves structural fidelity (GED, Clash) and diffusion enhances conformational exploration (RMSD, Clash). Best results are bold.} 
\label{table:ablation_model}
\begin{center}\small % Preserve original font size
\begin{sc} % Preserve original small caps style
    \begin{tabular}{lcccc}
    \toprule
    Method & RMSD ($\downarrow$) & GED ($\downarrow$) & Clash ($\downarrow$) & DIV ($\downarrow$) \\
    \midrule
    VAE+diff & 
    1.813\scriptsize{$\pm$0.108} & 
    0.829\scriptsize{$\pm$0.333} & 
    0.376\scriptsize{$\pm$0.049} & 
    0.407\scriptsize{$\pm$0.030} \\
    VQ-VAE+flow & 
    1.805\scriptsize{$\pm$0.113} & 
    0.491\scriptsize{$\pm$0.070} & 
    0.367\scriptsize{$\pm$0.048} & 
    \textbf{0.391\scriptsize{$\pm$0.033}} \\
    \rowcolor{gray!20} VQ-VAE+diff & 
    \textbf{1.710\scriptsize{$\pm$0.089}} & 
    \textbf{0.471\scriptsize{$\pm$0.077}} & 
    \textbf{0.325\scriptsize{$\pm$0.044}} & 
    0.452\scriptsize{$\pm$0.020} \\
    \bottomrule
    \end{tabular}
\end{sc}
\end{center}
\end{table}

We performed ablation studies on the PED dataset to assess key architectural components of our protein backmapping model. Our proposed approach (VQ-VAE with diffusion) was compared against a flow-based alternative (VQ-VAE with flow matching) and a continuous latent space variant (VAE with diffusion). While flow matching offers efficient probability path computation~\citep{irwin2024efficient, jing2024alphafold}, its suitability for exploring the vast conformational landscapes inherent in protein backmapping is less established than diffusion models.

First, the diffusion model demonstrates greater efficacy than flow matching (VQ-VAE+flow) for navigating large conformational spaces. Diffusion's stochastic exploration of diverse conformations while maintaining structural validity leads to superior RMSD and Clash metrics (Table~\ref{table:ablation_model}). Specifically, the multi-step denoising process in diffusion facilitates finer structural adjustments, reducing steric clashes and better preserving molecular interactions, as reflected by lower Clash values.

Second, our discrete latent space model (VQ-VAE) significantly outperforms its continuous counterpart (VAE) in structural fidelity. Continuous VAEs often struggle with mode collapse when representing diverse conformations of the same protein, potentially blurring distinct structural features. In contrast, the VQ-VAE's discretization mitigates this issue, enabling a clearer representation of varied conformational states. This, coupled with better preservation of bond graph consistency—vital for accurate internal coordinates—results in markedly lower GED scores (Table~\ref{table:ablation_model}) and thus improved structural precision.

%%%%%%%%%%%%%%%%%
%%%%%%%%%%%%%%%%%
% Relate work
%%%%%%%%%%%%%%%%%
%%%%%%%%%%%%%%%%%
\section{Related Work}

\subsection{Traditional Methods}
Traditional backmapping methods employ rule-based heuristics for initial structure generation, followed by refinement via geometric optimization or energy minimization~\citep{vickery2021cg2at2}. 
However, these approaches often yield non-physical structures with atomic clashes and abnormal bond angles; this refinement is often computationally costly and biased by the minimization scheme~\citep{badaczewska2020computational}. Additionally, these methods are deterministic and do not capture the thermodynamic diversity of atomic structures that correspond to a single CG representation~\citep{yang2023chemically}.

\subsection{Data-driven Methods}
Data-driven approaches address these limitations by predicting atomic structures from CG inputs. 
Deterministic models (e.g., MLPs~\citep{an2020machine}, SE(3)-Transformers~\citep{heo2023one}) offer precision but limited structural diversity due to backmapping's one-to-many nature.
For instance, Gaussian Mixture Models (GMMs) have been used for local rotamer states coupled with a global model for protein conformations~\citep{chennakesavalu2024data}; however, reliance on such physical/statistical frameworks and a lack of end-to-end optimization, unlike direct distribution learning, can compromise accuracy.

Generative models, such as GANs~\citep{li2020backmapping,stieffenhofer2020adversarial,stieffenhofer2021adversarial,shmilovich2022temporally} and VAEs~\citep{wang2019coarse,wang2022generative,yang2023chemically}, tackle these issues by learning multi-modal atomic structure distributions. \citep{stieffenhofer2021adversarial} showcased generalization potential by training on small molecules and applying them to corresponding polymer systems. 
\citep{shmilovich2022temporally} proposed a GAN-based method that uses previous all-atom structures as temporal references to constrain reverse mapping.
\citep{yang2023chemically} leveraged a conditional VAE (cVAE) to align CG and all-atom distributions, successfully generalizing to novel proteins on the PED dataset. 
Despite these advances, GANs often struggle with complex distributions, and VAEs are prone to mode collapse, hindering the generation of diverse conformations.

Recently, diffusion models have emerged as a promising avenue for backmapping, owing to their proficiency in generating diverse and high-fidelity molecular structures through stochastic sampling \citep{li2024towards,jones2023diamondback,liu2023backdiff,jones24flowback, zhang2025exploit}.
DiAMONDBack~\citep{jones2023diamondback} performs residue-by-residue denoising to restore atomic details, which improves local accuracy but substantially increases computational cost and compromises global geometric consistency. 
Backdiff~\citep{liu2023backdiff} constrains denoised structures by projecting them onto their corresponding CG configurations and penalizing deviations from these targets, while additionally enforcing bond-length and bond-angle consistency to ensure chemical and structural validity. Although such explicit constraint handling substantially increases the number of denoising steps—leading to lower sampling efficiency—it enables the model to generalize across diverse CG mappings within a unified diffusion framework.
FlowBack \citep{jones24flowback} performs denoising directly within the all-atom space using probability flow interpolation, yet it demonstrates restricted generalization capabilities across varied conformational systems.
Overall, for standard diffusion models tackling multi-conformational problems, intricate constraints vital for structural validity often impose adverse trade-offs in accuracy, efficiency, and generalization. 
CODLAD mitigates these challenges by performing diffusion in a compact all-atom latent space, implicitly preserving structural validity while substantially improving sampling efficiency.

%%%%%%%%%%%%%%%%%
%%%%%%%%%%%%%%%%%
% Conclusion
%%%%%%%%%%%%%%%%%
%%%%%%%%%%%%%%%%%
\section{Conclusion}

In this work, we present CODLAD, a novel two-stage latent diffusion framework for protein backmapping that decouples structural constraint handling from the generative process. 
By shifting constraint enforcement to a hierarchical compression stage, CODLAD enables unconstrained and efficient sampling in a low-dimensional latent space. 
This design effectively resolves the trade-off between maintaining atomistic accuracy and exploring diverse conformations.
Extensive experiments on static and dynamic protein datasets demonstrate that CODLAD consistently achieves state-of-the-art performance in atomistic accuracy, conformational diversity, computational efficiency, and generalization. 
Notably, CODLAD remains robust on challenging cases such as ATLAS and DES, where existing all-atom diffusion methods often fail to generalize or preserve structure. 
Compared to prior approaches that tightly couple constraint satisfaction with atom-level generation, CODLAD offers a principled decoupling that improves scalability without compromising quality.
We believe this framework establishes a new paradigm for efficient and accurate protein structure backmapping.

While CODLAD demonstrates strong generalization and efficiency, its current formulation is restricted to C$_\alpha$-based coarse-graining. Extending the method to alternative or mixed-resolution mappings will likely require retraining or adaptation of the latent diffusion model to accommodate different topologies and feature spaces. 
Future work will explore incorporating latent space energy-based priors or latent space lightweight force-field feedback to further enhance physical fidelity and transferability without reintroducing hard constraints in Cartesian space.
We anticipate that such latent space modeling will facilitate broader application of CODLAD across biomolecular systems of varying resolution.

\section*{Data and Code Availability}
The complete resources associated with this work, including the source code, datasets, and pre-trained model weights, are publicly available at \url{https://github.com/xiaoxiaokuye/CODLAD}.


\begin{suppinfo}
The Supporting Information is available free of charge at \url{https://pubs.acs.org/doi/10.1021/acs.jctc.XXXXXXX}.
\begin{itemize}
  \item Algorithms S1--S4 detailing the procedures for internal-to-Cartesian reconstruction, atom placement subroutines, model training, and inference strategies.
\end{itemize}
\end{suppinfo}

\section*{Notes}
The authors declare no competing financial interest.

\begin{acknowledgement}
This work was supported by the Administrative Committee of Zhongguancun Science City.
\end{acknowledgement}

\bibliography{refs}

\clearpage
\newpage

\begin{tocentry}
%\noter{Add TOC image and caption.}

% Constraint-decoupled backmapping: Stage (a) learns a constraint-consistent discrete latent of all-atom structures; Stage (b) performs CG-conditioned diffusion in latent space and decodes to valid AA coordinates.

%Some journals require a graphical entry for the Table of Contents.
%This should be laid out ``print ready'' so that the sizing of the

%Inside the \texttt{tocentry} environment, the font used is Helvetica
%8\,pt, as required by \emph{Journal of the American Chemical
%The surrounding frame is 9\,cm by 3.5\,cm, which is the maximum
%permitted for  \emph{Journal of the American Chemical Society}
%graphical table of content entries. The box will not resize if the
%content is too big: instead it will overflow the edge of the box.
%
%This box and the associated title will always be printed on a
%separate page at the end of the document.
\begin{center}
\includegraphics[width=2.5in]{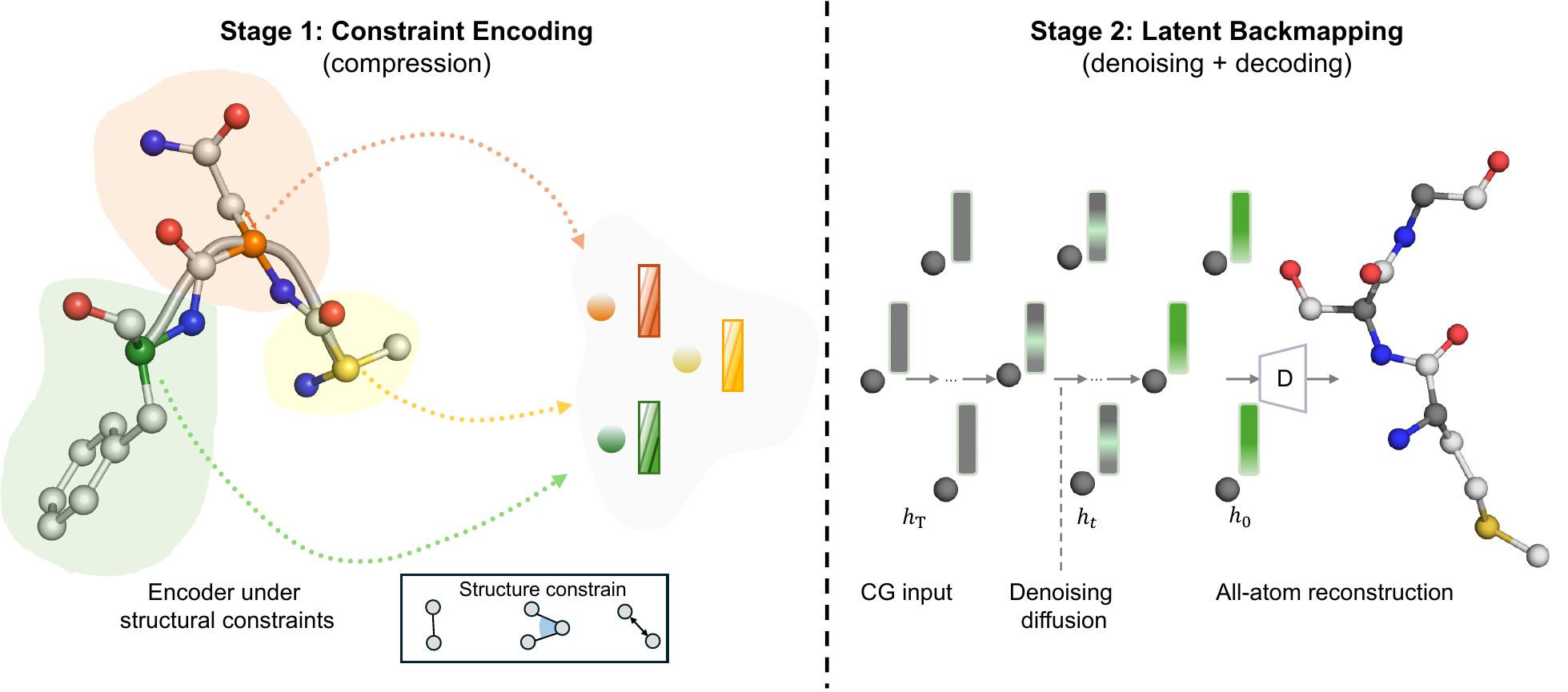}
\end{center}
\end{tocentry}

\newpage
\includepdf[pages=-]{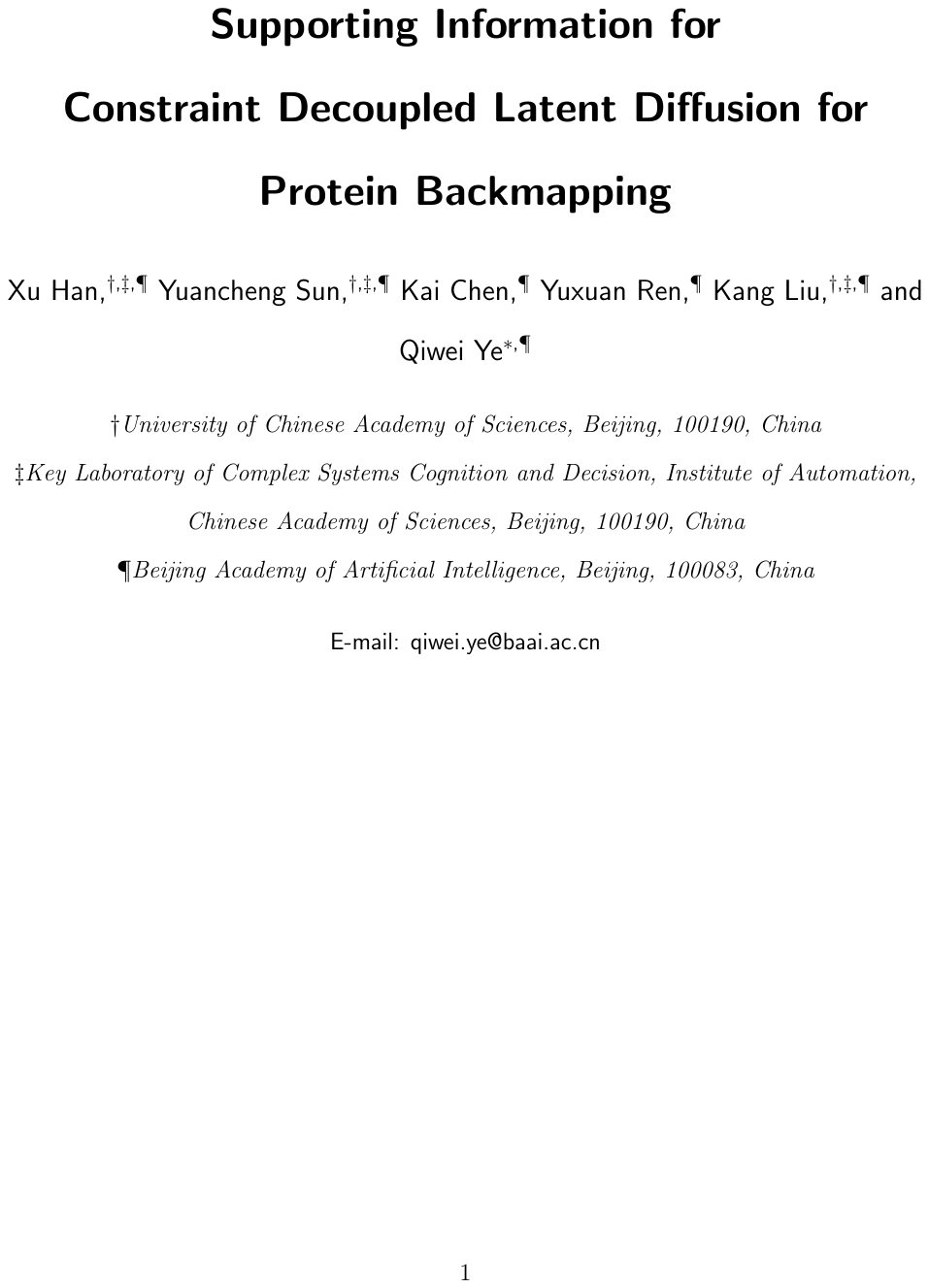}

\end{document}